\documentclass[10pt,twocolumn,letterpaper]{article}

\usepackage[pagenumbers]{wacv} 

\usepackage{graphicx}
\usepackage{amsmath}
\usepackage{amssymb}
\usepackage{booktabs}

\usepackage[pagebackref,breaklinks,colorlinks]{hyperref}

\usepackage{amsmath}
\usepackage{amssymb}
\usepackage{multirow}
\usepackage{bm}
\usepackage{caption}
\usepackage{algorithm}
\usepackage{algpseudocode}

\usepackage{pifont}
\newcommand{\cmark}{\ding{51}}%
\newcommand{\xmark}{\ding{55}}%
\usepackage{xr}

\newcommand\blfootnote[1]{
    \begingroup
    \renewcommand\thefootnote{}\footnote{#1}
    \addtocounter{footnote}{-1}
    \endgroup
}

\usepackage[capitalize]{cleveref}
\crefname{section}{Sec.}{Secs.}
\Crefname{section}{Section}{Sections}
\Crefname{table}{Table}{Tables}
\crefname{table}{Tab.}{Tabs.}

\begin{document}

\title{Towards Unsupervised Blind Face Restoration using Diffusion Prior}

\author{
    Tianshu Kuai$^{2,\dagger}$,
    Sina Honari$^{1}$, 
    Igor Gilitschenski$^{2,3}$, 
    Alex Levinshtein$^{1}$ \and
    $^{1}$Samsung AI Center Toronto, $^{2}$University of Toronto, $^{3}$Vector Institute for AI
}
\maketitle

\begin{abstract} 
Blind face restoration methods have shown remarkable performance, particularly when trained on large-scale synthetic datasets with supervised learning. These datasets are often generated by simulating low-quality face images with a handcrafted image degradation pipeline. The models trained on such synthetic degradations, however, cannot deal with inputs of unseen degradations. In this paper, we address this issue by using only a set of input images, with unknown degradations and without ground truth targets, to fine-tune a restoration model that learns to map them to clean and contextually consistent outputs. We utilize a pre-trained diffusion model as a generative prior through which we generate high quality images from the natural image distribution while maintaining the input image content through consistency constraints. These generated images are then used as pseudo targets to fine-tune a pre-trained restoration model. Unlike many recent approaches that employ diffusion models at test time, we only do so during training and thus maintain an efficient inference-time performance. Extensive experiments show that the proposed approach can consistently improve the perceptual quality of pre-trained blind face restoration models while maintaining great consistency with the input contents. Our best model also achieves the state-of-the-art results on both synthetic and real-world datasets. \href{https://dt-bfr.github.io/}{Project Page}. 
   \blfootnote{$^\dagger$Work done during an internship at Samsung AI Center Toronto.}
\end{abstract}

\section{Introduction}
\label{sec:intro}

\begin{figure}[t!]
    \centering
    \includegraphics[width=\linewidth, trim={0cm 0.0cm 0cm 0.0cm},clip]{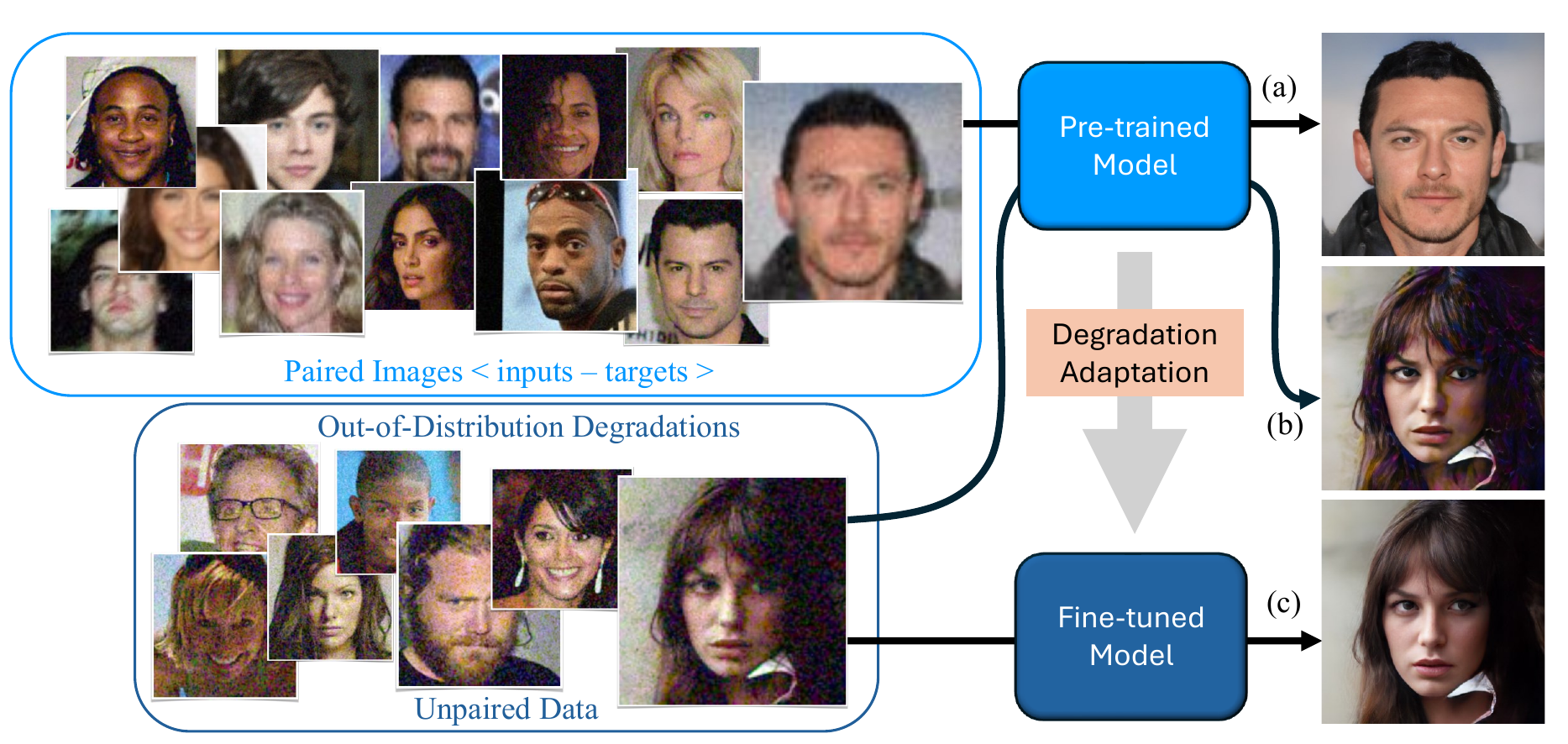}
    \caption{\textbf{Overview.} Given a restoration model pre-trained on synthetic datasets in a supervised fashion, it can produce high-quality restoration on low-quality images that are aligned with the degradation distribution used in training (a). However, it often fails on inputs of out-of-distribution degradations (b). We propose an unsupervised pipeline to adapt a pre-trained model to unpaired degraded images of the target degradation with a much smaller data size. This addresses the domain gap in degradation types without paired ground-truth images or the knowledge of the target data's degradation type (c). (\textbf{zoom in for details}).}
    \label{fig:teaser}
    \vspace{-10pt}
\end{figure}

Image restoration is a fundamental task in computational photography that aims to recover a high-quality image from its degraded low-quality counterpart. Blind image restoration is a more challenging task, where the degradation process is unknown. One needs to find a good balance between maintaining the fidelity of the image content and the output's perceptual quality. This is particularly important in the case of blind face restoration, as both fidelity and quality are important when restoring face images. 

Most of the existing blind face restoration methods~\cite{wang2022restoreformer,gfpgan,vqfr,codeformer,gpen} are trained in a supervised manner using a paired dataset of low-quality inputs and high-quality target images. The training pairs are often constructed by manually designing a degradation process~\cite{gfpgan,li2018learning,li2020blind}, where a high-quality image is synthetically degraded to form the corresponding low-quality input. Supervised learning achieves great performance on test data that aligns with the training degradations (\cref{fig:teaser}(a)). However, this produces results with severe artifacts when tested on images that do not fall under the training degradation distribution (\cref{fig:teaser}(b)). In addition, ground-truth data is not always available for the supervised learning setup. We commonly have access to only the low-quality observations in a real-world setting. In this paper, we address such blind unsupervised setup with neither access to the paired ground truth images nor to the degradation process of the inputs.

Image diffusion models~\cite{ho2020denoising,rombach2022high} have recently shown remarkable performance in image generation. Due to their powerful modeling of the natural image manifold, pre-trained diffusion models can be used as priors for image restoration tasks in a zero-shot manner. 
Some blind image restoration methods~\cite{wang2023dr2,yang2023pgdiff,yue2022difface, lin2023diffbir,ding2023restoration} have achieved great results on severely degraded data. However, they require running the diffusion model's sampling process during inference, which results in a significant computational cost and extremely slow runtime. 
On the other hand, a series of works~\cite{wang2022zero,kawar2022denoising,lugmayr2022repaint,fei2023generative, chung2022diffusion,luo2023image,zhu2023denoising} involve designing hand-crafted denoising process for known degradation types in a supervised setup, which limits the applicable scenarios. 

In this work, we tackle the problem of unsupervised blind image restoration by taking the advantages from the above two groups of works. Given a pre-trained restoration model that fails on inputs with some out-of-distribution degradations (target degradations), our approach consists of two stages: pseudo target generation and model fine-tuning. To generate pseudo targets, we design a denoising diffusion process that cleans up the restoration model's outputs, where it preserves the input image content while considerably enhancing high frequency details. In the second step, the cleaned images are treated as pseudo targets to fine-tune the pre-trained restoration model using input and pseudo target pairs. The fine-tuned model is then able to handle inputs with the same target degradations (\cref{fig:teaser}(c)). The proposed approach requires only a small set of unpaired low-quality observations for training, and does not require running the diffusion model at test time, which is a much more practical setting for real-world applications. To the best of our knowledge, this is the first approach to use pseudo-targets to adapt a pre-trained restoration model to unknown degradations for blind face restoration. 

In summary, our contributions are as follows: (i) 
An unsupervised pipeline for adaptation of face restoration models to unknown unpaired degradations; (ii) a method to obtain content-preserving pseudo targets from a diffusion model that achieves better fidelity and perceptual quality than previous zero-shot diffusion-based restoration methods; (iii) our approach consistently improves the pre-trained models, and our best fine-tuned model achieves state-of-the-art performance on both synthetic and real-world datasets without the need for running a diffusion model at inference time.

\section{Related Work}
\label{sec:related_work}

\begin{figure*}[t]
\begin{center}
\includegraphics[width=1.0\linewidth]{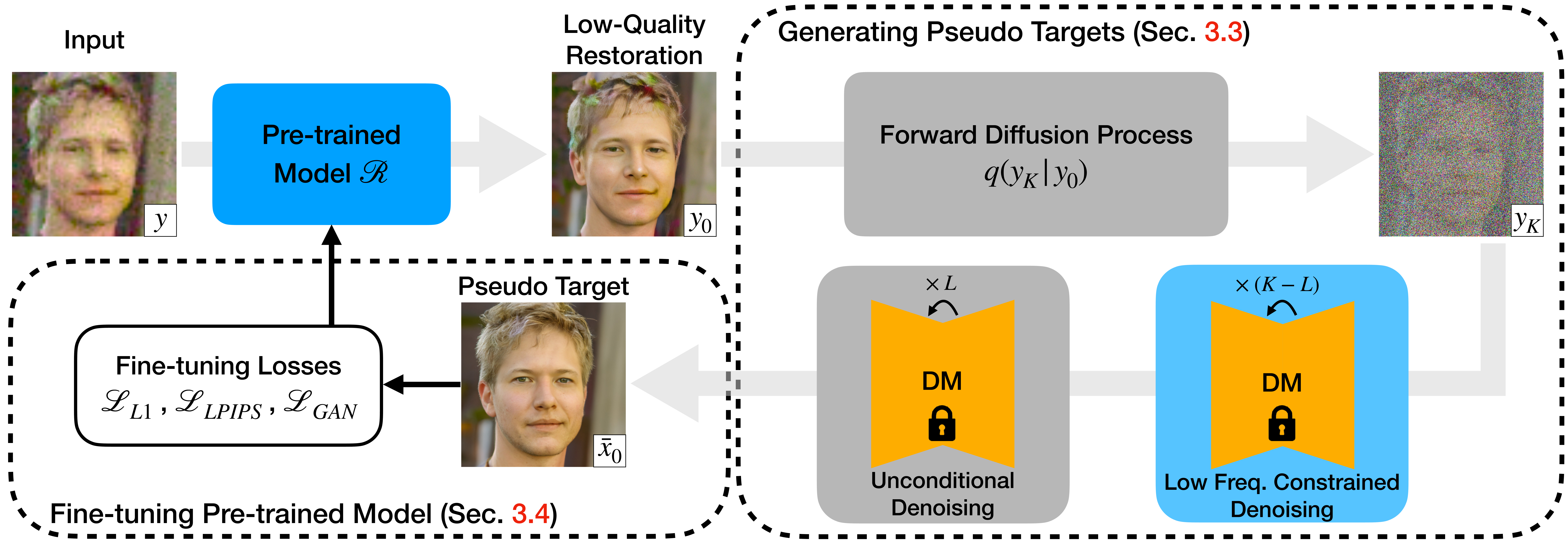}
\end{center}
\vspace{-10pt}
\caption{\textbf{Overview of our unsupervised fine-tuning pipeline.} Given a pre-trained restoration model that produces low-quality restoration outputs (severe artifacts on hair and over-smoothed skin) on samples with unknown and out-of-distribution degradations, we generate pseudo targets using a pre-trained unconditional diffusion model with a combination of low frequency content constrained denoising and unconditional denoising. The generated clean images can be used as pseudo GT to fine-tune the pre-trained restoration model without the need for real GT images. }
\label{fig:overview}
\vspace{-1.0em}
\end{figure*}

\subsection{Supervised Blind Face Restoration} 

Most of the existing methods in blind face restoration involve supervised learning with simulated training data pairs, combined with various types of priors. Due to the structured nature of facial images, many works explore face-specific priors such as geometry priors~\cite{chen2018fsrnet,kim2019progressive,zhu2016deep,shen2018deep,chen2021progressive,yu2018face,zhu2022blind,yasarla2020deblurring} and reference priors~\cite{li2018learning,li2020enhanced,dogan2019exemplar,li2022learning} to retain natural and faithful restoration for the given low-quality faces. To further improve the perceptual quality of the restoration, several models~\cite{gpen,gfpgan,glean,pulse,li2023analyzing} employ GAN-based priors with perceptual and adversarial losses during supervised training, or use pre-trained GAN models~\cite{ffhq,karras2020analyzing} as priors directly. Some works~\cite{li2022learning,li2020blind} explore facial component dictionaries as a more robust prior for higher quality restoration and identity preservation.

Following the high-quality codebook learning approaches~\cite{esser2021taming,van2017neural}, more recent methods~\cite{zhao2022rethinking,wang2022restoreformer,vqfr,codeformer} show great performance on real-world degraded faces by first pre-training on large-scale clean face data to obtain high-quality discrete dictionaries or codebooks as priors during restoration. These methods achieve great perceptual quality, but do not generalize to the degradations that are outside of the training degradation distribution. The reliance on the large-scale training data pairs also limit their practicality for real-world applications. Different from the previous supervised approaches, we aim to address the generalization problem of pre-trained face restoration models on out-of-distribution degradations in an unsupervised manner.

\subsection{Diffusion Priors for Image Restoration} Diffusion models~\cite{ho2020denoising,rombach2022high} have recently become the most powerful generative models in image synthesis. Despite not being initially designed for low-level imaging tasks, some works modify diffusion model's architecture and train them by conditioning on either the degraded image or its features for tasks such as super-resolution (SR)~\cite{gao2023implicit, sahak2023denoising, sr3, luo2023refusion, xia2023diffir}, shadow-removal~\cite{guo2023shadowdiffusion, luo2023refusion}, deblurring~\cite{ren2023multiscale, xia2023diffir}, inpainting and uncropping~\cite{saharia2022palette, xia2023diffir}, face-restoration~\cite{zhao2023towards}, and adverse weather
restoration~\cite{ozdenizci2023restoring, luo2023refusion}. Some works~\cite{whang2022deblurring,luo2023image,qiu2023diffbfr} explore different sampling and training procedure of diffusion models for better restoration performance. However, all of these models require supervised training with large amount of data pairs and computational resources, while still suffering from lack of adaptability to generalize to out-of-distribution degradations. 

A large group of the methods utilize pre-trained diffusion models for zero-shot image restoration tasks 
~\cite{kawar2022denoising,wang2022zero,fei2023generative,zhu2023denoising,song2022pseudoinverse,chung2022diffusion,kawar2021snips,murata2023gibbsddrm,kawar2022jpeg,rout2024solving,chung2023parallel,chung2022come,dou2023diffusion,feng2023score} 
including super-resolution, inpainting, deblurring, denoising, and JPEG artifact correction. However, they require the knowledge of the degradation type to design custom denoising process, which cannot be applied to blind image restoration directly. To tackle the blind restoration problem, where the degradation is unknown, the conditioning or guidance during the denoising process needs to be generalized enough to handle a variety of degradation types. On this line of research, some papers~\cite{ilvr,gao2023back,wang2023dr2} use the low frequency content of the images to guide the denoising process in order to preserve the content of the image. Some other papers~\cite{yue2022difface,lin2023diffbir,yang2023pgdiff} use a simple pre-trained restoration model to first reduce the amount of artifacts in the image while preserving the smoothed semantics, and then use a pre-trained or fine-tuned diffusion model to inject sharp textures to the restoration model's output. Although these zero-shot approaches achieve great level of perceptual quality, they share the common problem of long inference time due to running the diffusion model for every input. There have been efforts to make the diffusion models faster by reducing the number of sampling using DDIM~\cite{song2020denoising} or progressively distilling it into fewer steps~\cite{salimans2022progressive} at the expense of image quality. Our approach elimiates the burden of running the diffusion model altogether at inference time. It only uses the diffusion model's outputs during training as pseudo targets to improve a pre-trained restoration model by injecting its prior to  restore unknown degradations.
\section{Method}
\label{sec:method}

\begin{figure*}[ht]
\begin{center}
\includegraphics[width=1.0\linewidth]{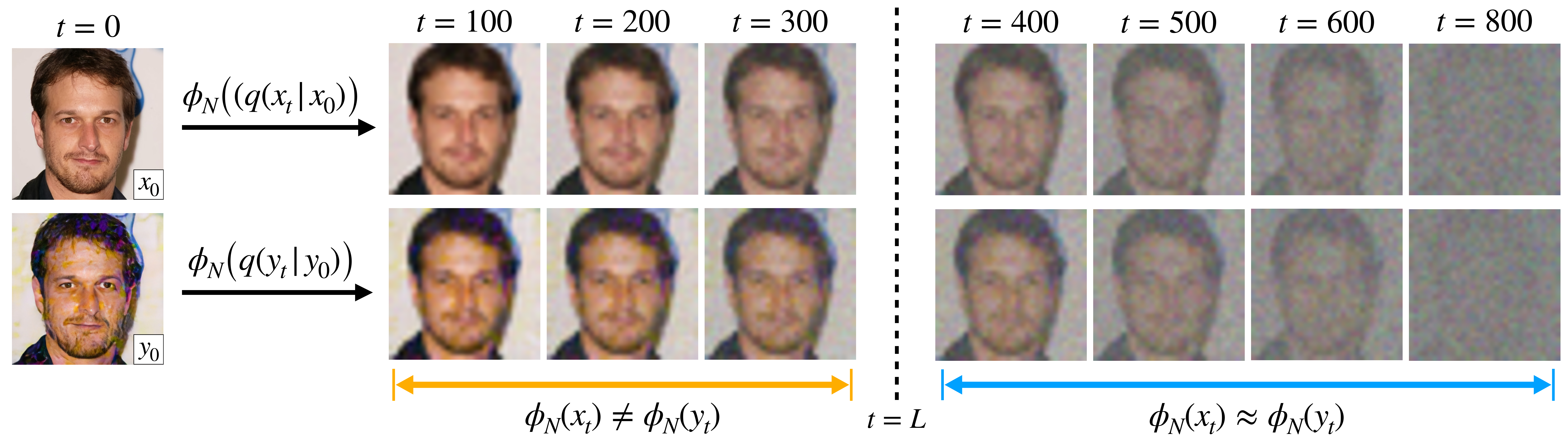}
\end{center}
\vspace{-10pt}
\caption{\textbf{Visualization of low frequency contents at different timesteps.} We show low frequency contents of the low-quality restoration from a pre-trained SwinIR~\cite{liang2021swinir} and its GT counterparts at different timesteps of the forward diffusion process (\textbf{zoom in for details}).}
\label{fig:low_freq_vis}
\vspace{-1.0em}
\end{figure*}

\subsection{Preliminaries on Diffusion Models}

Denoising Diffusion Probabilistic Model (DDPM)~\cite{sohl2015deep, ho2020denoising} has been one of the most used and powerful generative models in computer vision. An unconditional diffusion model learns a natural image manifold from large-scale image datasets. It follows a Markov forward process to gradually corrupt an image ${x_0}$ with a pre-defined Gaussian noise variance schedule ${\beta_t}$ for each timestep $t\in\{1,2,...,T\}$. Thanks to the Markovian formulation of the diffusion process, one can write the expression for the noisy image $x_t$ at any timestep $t$, given the clean image $x_0$ as
\begin{align}
    q(x_t | x_0) &= \mathcal{N}(x_t; \sqrt{\bar \alpha_t}x_{0},   (1 - \bar \alpha_t)\boldsymbol{\mathrm{I}}),
    \label{eq:forward}
\end{align}
where ${\alpha}_t = 1 - \beta_t$ and $\bar{\alpha}_t = \prod_{s=1}^t \alpha_s$. 

An unconditional diffusion model generates natural images by reversing the forward diffusion process. Specifically, the process can be written as:
\begin{equation}
    p_\theta(x_{t-1} | x_t) = \mathcal{N}(x_{t-1}; \boldsymbol{\mu}_\theta(x_t, t), \sigma_t^2 \boldsymbol{\mathrm{I}}),
    \label{eq:reverse}
\end{equation}
where $\sigma_t^2$ is set to a time-dependent constant, and the mean of the denoised image $\boldsymbol{\mu}_\theta(x_t, t)$ is computed as:
\begin{equation}
    \boldsymbol{\mu}_\theta(x_t, t) = \frac{1}{\sqrt{\alpha_t}}\Bigl(x_t - \frac{1 - \alpha_t}{\sqrt{1 - \bar{\alpha}_t}} \boldsymbol{\epsilon}_\theta(x_t, t)\Bigr),
    \label{eq:network}
\end{equation}
where the noise $\boldsymbol{\epsilon}_\theta(x_t, t)$ at timestep $t$ is predicted by a trained timestep conditioned U-Net~\cite{unet}. To perform unconditional image generation, we can start with a sample from the standard Gaussian distribution $x_T \sim \mathcal{N}(\bm{0}, \boldsymbol{I})$ and gradually denoise it using the predicted noise at each timestep. Note that one can use techniques from DDIM~\cite{song2020denoising} or simply uniformly skip timesteps during the reverse diffusion process to accelerate the denoising process.

\subsection{Method Overview}
Given a restoration model that is pre-trained on synthetic data pairs, its performance on out-of-distribution inputs is often heavily degraded. To bridge the gap without the need for paired ground-truth high-quality images, we use a pre-trained unconditional diffusion model~\cite{ho2020denoising} to clean up the artifacts in restoration model's output via a low frequency constrained denoising process. As shown in~\cref{fig:overview}, our pipeline consists of two stages: 1) generating pseudo targets using a diffusion model (Section~\ref{sec:targets}) and 2) using the generated targets to fine-tune the pre-trained restoration model (Section~\ref{sec:finetuning}).

\subsection{Generating Pseudo Targets}
\label{sec:targets}

Consider a pre-trained restoration model, $\mathcal{R}$, and a real-world low-quality image observation $y$. Due to the domain gap between the synthetic data and the real-world data, the output from a pre-trained restoration model, $y_{0}=\mathcal{R}(y)$, still contains a lot of artifacts. Following~\cref{eq:forward}, we can apply the forward diffusion process on $y_0$, to get a noisy version of the low-quality restoration $y_t$ at timestep $t$ as:
\begin{equation}
    y_t = \sqrt{\bar \alpha_t}y_0 + \sqrt{1-\bar\alpha_t}\boldsymbol{\epsilon}, \text{ where } \boldsymbol{\epsilon} \sim \mathcal{N}(\mathbf{0}, \boldsymbol{\mathrm{I}}).
    \label{eq:forward_diffusion_lq}
\end{equation}
If we directly perform an unconditional denoising on the noisy image $y_t$ as in~\cite{yue2022difface}, the structured content will not be preserved well, which yields inconsistent restoration. 
Similar to the observations from~\cite{wang2023dr2,ding2023restoration}, we found that as more Gaussian noise (larger timestep) is added to the low-quality restoration, the low frequency content of $y_t$ is getting closer to the low frequency content of the noisy image $x_t$ as if we start with the clean image counterpart $x_0$. Specifically, given a low pass filter $\phi_{N}$ and if $t$ is large enough ($t > L$):
\begin{equation}
    \phi_N(y_t) \approx \phi_N(x_t) = \phi_N(\sqrt{\bar \alpha_t}x_0 + \sqrt{1-\bar\alpha_t}\boldsymbol{\epsilon}),
    \label{eq:forward_diffusion_clean}
\end{equation}
where $x_t$ is the noisy version of the clean image, and $\boldsymbol{\epsilon}$ is the same sampled noise in~\cref{eq:forward_diffusion_lq}. We visualize and compare the low frequency contents of the two images at different timesteps in~\cref{fig:low_freq_vis}. The noisy images become closer as timestep increases, and eventually indistinguishable visually after $t=400$. With this critical observation, we can constrain the denoising process by regularizing the low frequency content at each denoising timestep when $t > L$, in order to preserve the structural information. At lower timesteps ($t <= L$), the low frequency property in Eq.~\ref{eq:forward_diffusion_clean} no longer holds and applying such regularization will deteriorate the denoising process. In addition, since we are using an unconditional diffusion model, going all the way to $t=T$ would completely destroy all the information in the image. Hence, we start the low frequency constrained denoising process at a smaller timestep $t=K$, where the low frequency content is not yet destroyed by the injected Gaussian noise.

Combined with the insights above, we describe our pseudo target generation process as follows: we take the restoration model's output, $y_0$, on an image from the target dataset, and follow the pre-defined noise schedule to inject Gaussian noise into the image up to timestep $K$. We then pass it to the diffusion model to clean up the image. Similar to~\cite{wang2023dr2,ilvr,gao2023back}, we guide the denoising process by constraining the low frequency content to be consistent with the input. This is done by replacing the low frequency content of the denoised image with the corresponding part from the noisy copy of the input image at each time step. Some methods~\cite{ilvr,gao2023back} apply such guidance on all denoising steps, which would lead to blurry outputs with artifacts due to over-constraining the denoised images on information that can be a mixture of signal and noise.

\begin{figure}[t]
\begin{algorithm}[H]
\caption{Generating pseudo targets using a pre-trained diffusion model}
\begin{algorithmic}
\State \textbf{Input}: low-quality restoration output $y_0=\mathcal{R}(y)$, low-pass filter $\phi_{N}$
\State \textbf{Output}: pseudo target $\bar{x}_{0}$ for low-quality input $y$
\State $\bar{x}_K \gets \text{sample from}~\mathcal{N}(y_K; \sqrt{\bar \alpha_K}y_{0}, (1-\bar \alpha_K)\boldsymbol{\mathrm{I}})$
\For{$t$ from $K$ to $1$}
    \State $\bar{x}_{t-1} \gets \text{sample from } p_\theta(\bar{x}_{t-1} | \bar{x}_t)$ \Comment{unconditional denoising}
    \If{$ t > L$}
        \State ${y}_{t-1} \gets \text{sample from}~\mathcal{N}({y}_{t-1}; \sqrt{\bar \alpha_{t-1}}y_{0},   (1 - \bar \alpha_{t-1})\boldsymbol{\mathrm{I}})$
        \State $\bar{x}_{t-1} \gets \bar{x}_{t-1} - \phi_N(\bar{x}_{t-1}) + \phi_N({y}_{t-1})$ \Comment{low frequency content constraint}
     \EndIf
    \EndFor
\State \Return $\bar{x}_{0}$
\end{algorithmic}
\label{alg:sampling}
\end{algorithm}
\vspace{-2.0 em}
\end{figure}

Different from previous methods, we only apply this low frequency content constraint for timesteps when $t > L$. This is because the low frequency property does not hold anymore for small timesteps ($t \leq L$) as shown in~\cref{fig:low_freq_vis}. Moreover, we observe that the denoised images at these timesteps already have reasonably good structure. Therefore, we perform unconditional denoising for the remaining $L$ timesteps since unconditional denoising steps contribute to high-frequency details at small timesteps~\cite{meng2021sdedit}. With this approach, there is no need for directly estimating $x_0$ from $x_L$ in one step and running another enhancement model on the generated image~\cite{wang2023dr2}. We summarize the detailed procedure of generating pseudo targets in Algorithm~\ref{alg:sampling}.

\subsection{Fine-tuning the Pre-trained Models}
\label{sec:finetuning}

After obtaining the pseudo targets from the diffusion model, one can fine-tune the pre-trained restoration model with the low-quality inputs and pseudo targets data pairs. We apply the widely used image-level $L1$ loss, perceptual (LPIPS) loss~\cite{lpips}, and adversarial (GAN) loss~\cite{goodfellow2014generative}: 
\begin{align}
    & \mathcal{L}_{L1} = ||\mathcal{R}(y) - \bar{x}_0 ||_{1}, \quad \mathcal{L}_{LPIPS} = \text{LPIPS}\bigl(\mathcal{R}(y), \bar{x}_0\bigl), & \notag\\
    & \qquad \qquad \qquad \mathcal{L}_{GAN} = \text{log}\bigl(1 - \mathcal{D}(\mathcal{R}(y))\bigr), &
\label{Eq:finetune_loss_definitions}
\end{align}
where $\mathcal{R}$ is our restoration model, and $\mathcal{D}$ is a discriminator that outputs the probability of its input coming from the distribution of real natural faces. This discriminator is optimized from scratch along with the restoration model, with the following cross-entropy training objective~\cite{goodfellow2014generative}:
\begin{equation}
    \mathcal{L}_{\mathcal{D}} = \mathbb{E}_{x \sim \mathcal{R}(y)}\bigl[ - (1 - \text{log} \mathcal{D}(x))\bigr] + \mathbb{E}_{x \sim \mathbb{P}_r}\bigl[- \text{log} \mathcal{D}(x)\bigr],
    \label{Eq:discriminator}
\end{equation}
where $\mathbb{P}_r$ represents the distribution of real high-quality face images. In practice we treat the images in FFHQ dataset~\cite{ffhq} as our real data distribution. Therefore, we use randomly sampled images from the FFHQ dataset as clean references for optimizing the discriminator. This ensures that the discriminator is robust enough to provide useful gradient signals for optimizing the restoration model.
The complete training objective is as follows:
\begin{equation}
    \mathcal{L} = \mathcal{L}_{L1} + \lambda_{LPIPS}\mathcal{L}_{LPIPS} + \lambda_{GAN}\mathcal{L}_{GAN},
    \label{eq:finetune_loss}
\end{equation}
where $\lambda_{LPIPS}$ and $\lambda_{GAN}$ are hyperparameters for the weights of the losses.

\section{Experiments}
\label{sec:experiments}

\begin{figure*}[ht]
\begin{center}
\includegraphics[width=1.0\linewidth]{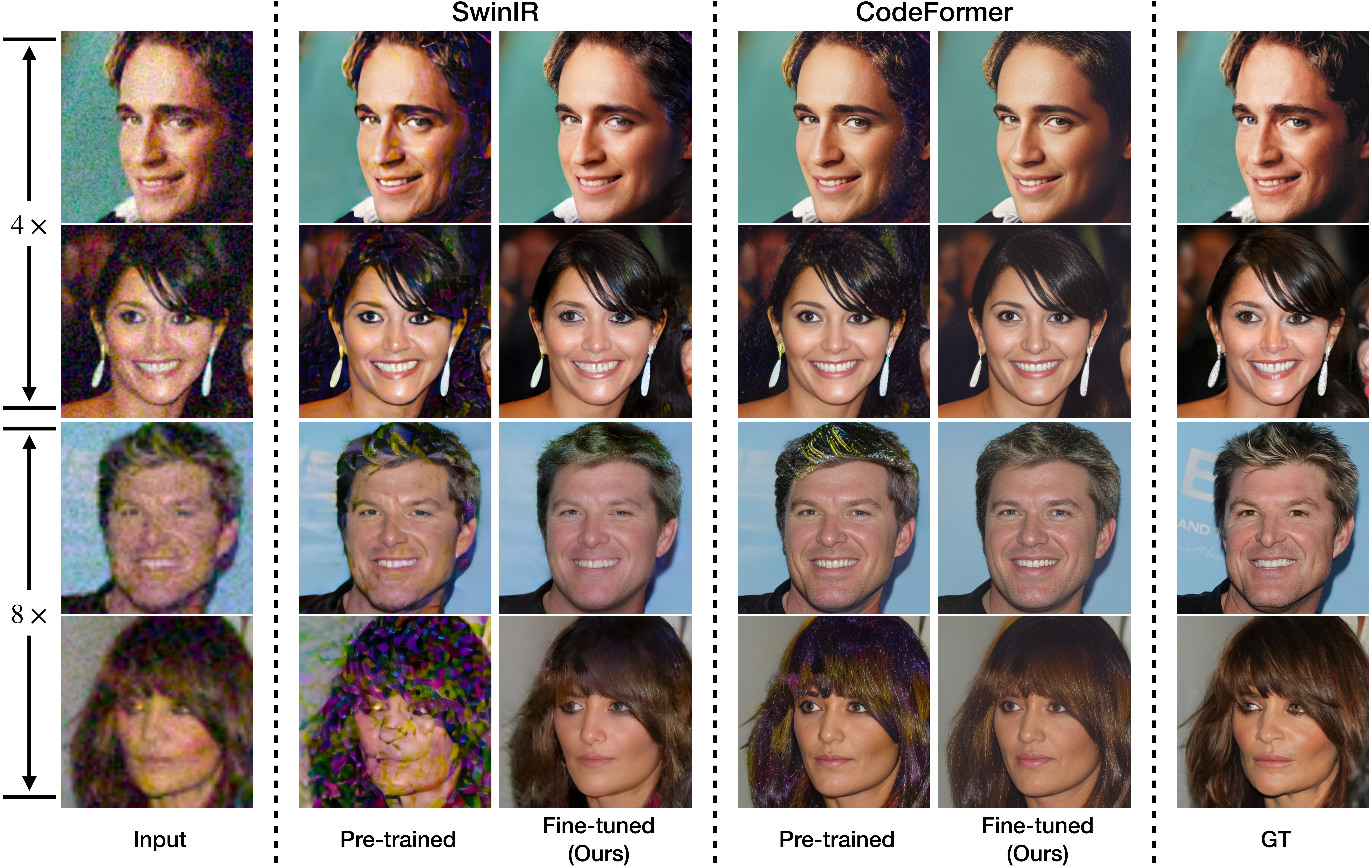}
\end{center}
\vspace{-10pt}
\caption{\textbf{Qualitative comparison of pre-trained versus fine-tuned models.} We show the effectiveness of our proposed approach to a pre-trained SwinIR~\cite{liang2021swinir} and a pre-trained CodeFormer~\cite{codeformer} models on 4$\times$ and 8$\times$ downsampled data at \textit{moderate} noise level. The fine-tuned models are able to produce realistic restoration (\textbf{zoom in for details}).}
\label{fig:finetuning_synthetic}
\end{figure*}

\begin{table*}[t!]
  \centering
  \resizebox{2.097\columnwidth}{!}{
      \begin{tabular}{l|c c c c|c c c c}
        \toprule
         & \multicolumn{4}{c|}{4$\times$ Downsampling} & \multicolumn{4}{c}{8$\times$ Downsampling} \\
          & PSNR$\uparrow$ & SSIM$\uparrow$ & LPIPS$\downarrow$ & FID$\downarrow$ & PSNR$\uparrow$ & SSIM$\uparrow$ & LPIPS$\downarrow$ & FID$\downarrow$ \\
        \midrule
         SwinIR~\cite{liang2021swinir}    & 21.28 / 20.92 & 0.5744 / 0.5444 & 0.5446 / 0.5842 & 74.12 / 120.44 & 21.28 / 19.00 & 0.5744 / 0.4813 & 0.5446 / 0.6435 & 99.44 / 152.90 \\
         SwinIR + Ours        & \textbf{24.75} / \textbf{23.41} & \textbf{0.6676} / \textbf{0.6284} & \textbf{0.3853} / \textbf{0.4156} & \textbf{41.42} / \textbf{49.46} & \textbf{23.28} / \textbf{21.91} & \textbf{0.6206} / \textbf{0.5720} & \textbf{0.4348} / \textbf{0.4821} & \textbf{68.25} / \textbf{115.97} \\
        \midrule
        \midrule
         CodeFormer~\cite{codeformer}    & \textbf{24.31} / \textbf{22.90} & \textbf{0.6335} / 0.5810 & \textbf{0.4007} / 0.4420 & \textbf{40.66} / 53.02 & 22.19 / 20.60 & 0.5716 / 0.5108 & 0.4420 / 0.4938 & 51.91 / 72.16 \\
         CodeFormer + Ours      & 23.20 / 22.85  & 0.6138 / \textbf{0.6036} & 0.4117 / \textbf{0.4258} & 41.74 / \textbf{41.21} & \textbf{22.28} / \textbf{21.38} & \textbf{0.5848} / \textbf{0.5514} & \textbf{0.4290} / \textbf{0.4589}  & \textbf{41.72} / \textbf{46.66}  \\
         \bottomrule
      \end{tabular}
    }
  \vspace{-5pt}
  \caption{Improvements gained on at 4$\times$ downsampling and 8$\times$ downsampling test sets at \textit{moderate} and \textit{severe} noise levels for pre-trained SwinIR~\cite{liang2021swinir} and CodeFormer~\cite{codeformer}. Numbers presented as: [\textit{moderate} / \textit{severe}].}
  \label{tab:finetuning_results}
  \vspace{-0.75 em}
\end{table*}

\subsection{Implementation and Evaluation Settings}
\label{sec:implementation_details}

\noindent \textbf{Pre-trained Models.}
We use a pre-trained unconditional face diffusion model with image resolution of 512$\times$512. This diffusion model is trained by following the training procedure from~\cite{guided, yue2022difface} on the entire FFHQ dataset~\cite{ffhq} with 70,000 images. We demonstrate the effectiveness of our approach on two model architectures: SwinIR~\cite{liang2021swinir} and CodeFormer~\cite{codeformer} (current state-of-the-art non-diffusion based method for blind face restoration). For SwinIR, we follow training setup from~\cite{gfpgan} to pre-train the SwinIR model on the FFHQ~\cite{ffhq} dataset following the losses in~\cref{eq:finetune_loss}. For CodeFormer, we use the pre-trained checkpoint from the authors. Both models are trained using the synthetic degradation function as in~\cite{gfpgan,li2018learning,li2020blind}:
\begin{equation}
    \mathcal{I}_{LQ} = \Bigl\{\bigl[(\mathcal{I}_{HQ} * \mathbf{k}_{\sigma})_{\downarrow_{r}} + \mathbf{n}_{\delta}\bigr]_{JPEG_{q}}\Bigr\}_{\uparrow_{r}},
    \label{Eq:common_degradation}
\end{equation}
where the high-quality image $\mathcal{I}_{HQ}$ is first convolved with a Gaussian blur kernel $\mathbf{k}_{\sigma}$ of kernel size $\sigma$ and downsampled by factor of $r$. Then Gaussian noise of standard deviation $\delta$ is added, followed by JPEG compression of quality factor $q$ and upsampling by a factor of $r$ to obtain the low-quality image $I_{LQ}$.

\noindent \textbf{Low Pass Filter and Timesteps.} We adopt the low pass filter from~\cite{ilvr}, where it is implemented as a sequence of downsampling and upsampling with factor of $N$. For the SwinIR pseudo targets, we set $N$ to be 16. For the CodeFormer pseudo targets, we set $N$ to be 8 for all the 4$\times$ data setup, and to be 16 for all the 8$\times$. We set the starting timestep to be $K=600$ and apply the low frequency constrained denoising process for $240$ timesteps. As a result, we start the unconditional denoising at the timestep of $L=360$. 

\noindent \textbf{Fine-tuning Setup.}
For fine-tuning the SwinIR model, we set the weights of the losses to be $\lambda_{LPIPS}=0.1$ and $\lambda_{GAN}=0.1$ for all the experiments. For CodeFormer ~\cite{codeformer}, we follow their training setup and empirically found that only adopting their code-level losses to optimize the code prediction module and the VQ-GAN encoder gives better fine-tuning performance than the image-level losses in~\cref{eq:finetune_loss}. Note that this fine-tuning procedure is specific to CodeFormer architecture and cannot be generalized to all model architectures. Please refer to the supplementary material for more details on our fine-tuning experiments.

\noindent \textbf{Testing Datasets.}
We evaluate our pipeline on both synthetic and real-world datasets. For the synthetic dataset, we generate low-quality testing inputs using 3,000 high-quality face images from the CelebA-HQ~\cite{celebahq} dataset. To simulate more realistic degradations instead of the the commonly used pipeline (\cref{Eq:common_degradation}) in~\cite{gfpgan,li2018learning,li2020blind}, we apply the following degradation model:
\begin{equation}
    \mathcal{I}_{LQ} = \Bigl\{ISP\bigl[ISP^{-1}((\mathcal{I}_{HQ})_{\downarrow_{r}}) + \mathbf{n}_{c}\bigr]\Bigr\}_{\uparrow_{r}},
    \label{Eq:realistic_degradation}
\end{equation}
where we first unprocess ($ISP^{-1}$) the downsampled high-quality images $\mathcal{I}_{HQ}$ to RAW~\cite{seo2023graphics2raw}, and then apply the widely used camera noise models~\cite{foi2008practical,liu2014practical,makitalo2012optimal} $\mathbf{n}_{c}$ to simulate noisy RAW images. We then render ($ISP$) the images back to sRGB~\cite{seo2023graphics2raw} and upsample the images to the same resolution as the high-quality ones. We construct datasets at \textbf{4}$\times$ and \textbf{8}$\times$ downsampling factors $\downarrow_{r}$. For each downsampling factor, we generate degraded images at \textit{moderate} and \textit{severe} noise levels. This yield four sets with each set containing 3,000 input-GT pairs. We use 2,500 images to generate the pseudo targets for fine-tuning, and use the remaining 500 pairs for evaluation. Note that during fine-tuning, we do not use the GT images and each set is fine-tuned independently. Due to limited space, we average the results on the  \textit{moderate} and \textit{severe} settings when presenting results for each one of \textbf{4}$\times$ and \textbf{8}$\times$ downsampling factors. More details on our dataset synthesis pipeline and the detailed results on each one of the four sets are provided in the supplementary material. For the real-world dataset, we use the Wider-Test set from~\cite{codeformer} for fine-tuning the pre-trained model, and we select another 200 severely degraded images from the testing set of the Wider-Face dataset~\cite{wider} as \textbf{Wider-Test-200} for evaluating fine-tuned models. Note that our Wider-Test-200 has no overlap with the set we obtained from~\cite{codeformer}.

\noindent \textbf{Evaluation Metrics.}
For synthetic datasets, we report PSNR, SSIM~\cite{ssim}, and LPIPS~\cite{lpips}, as we have access to the ground-truth images. For our Wider-Test-200 set, we report the non-reference image quality metrics (MANIQA~\cite{yang2022maniqa} and MUSIQ~\cite{ke2021musiq}). In addition, we report the commonly used FID scores~\cite{fid} for all datasets, where we use the distribution of the ground-truth images as the reference statistics for synthetic dataset. Since the Wider-Test-200 does not contain the ground-truth images, we use statistics of the FFHQ dataset~\cite{ffhq} as reference to measure the FID score.

\begin{figure*}[ht]
\begin{center}
\includegraphics[width=1.0\linewidth]{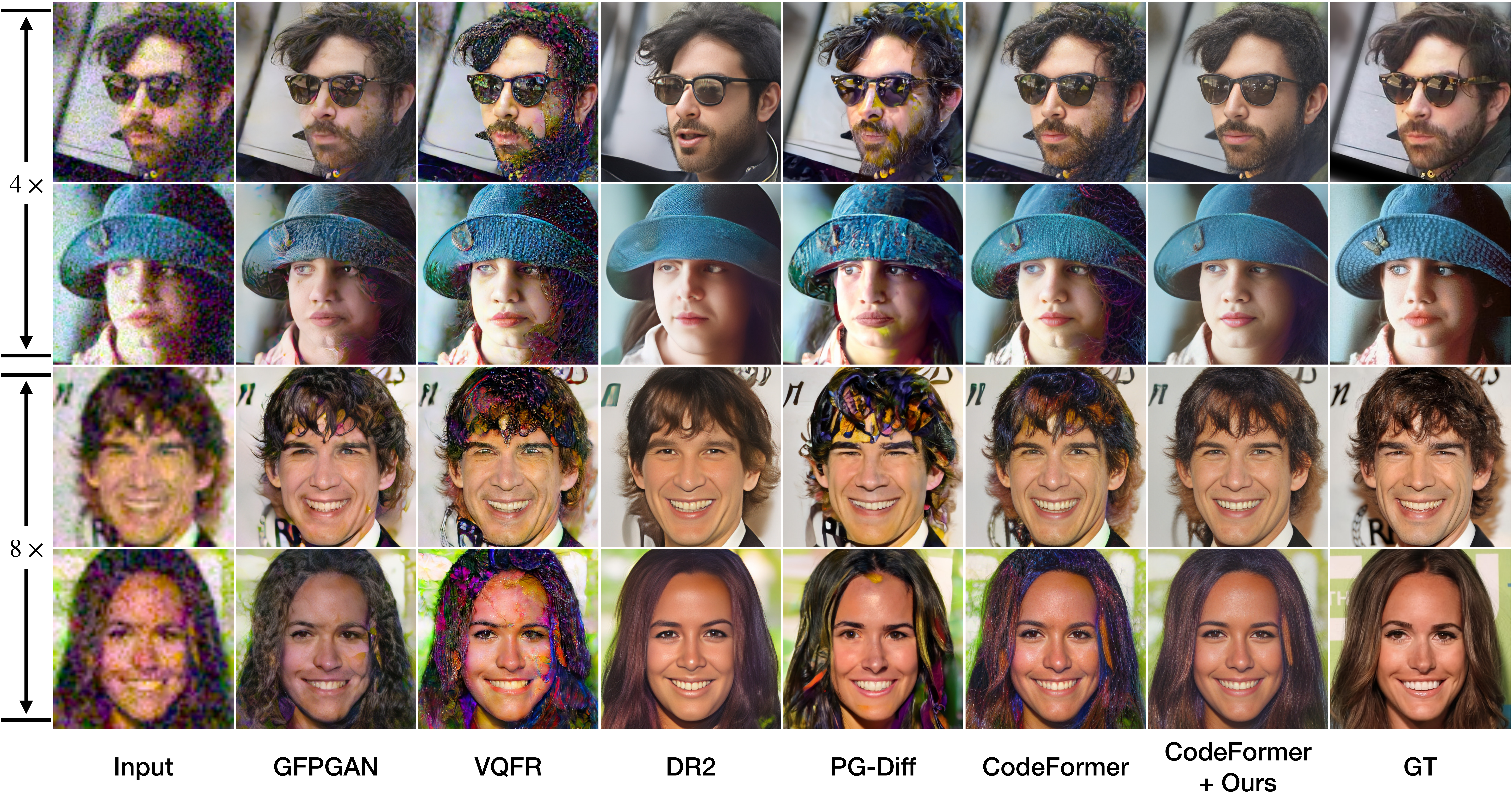}
\end{center}
\vspace{-15pt}
\caption{\textbf{Qualitative comparison with SOTA baselines on synthetic datasets.} Our fine-tuned CodeFormer model outperforms all other baselines and its pre-trained counterparts on severely degraded inputs from both 4$\times$ downsampling and 8$\times$ downsampling inputs (\textbf{zoom in for details}). }
\label{fig:result_synthetic}
\vspace{-1.0em}
\end{figure*}

\subsection{Results}
\label{sec:results}

\noindent \textbf{Pre-trained vs. fine-tuned models.}
We first compare the performance of the pre-trained models with the fine-tuned ones (obtained following our proposed approach in Section~\ref{sec:method}). We show the qualitative comparison in~\cref{fig:finetuning_synthetic}. 

For SwinIR~\cite{liang2021swinir}, the pre-trained model produces lots of artifacts. We achieve significant improvements on the perceptual quality of the restoration after fine-tuning.  We observe that the pre-trained CodeFormer~\cite{codeformer} already outputs faces in relatively good quality at lower degradation levels, but still produces noticeable artifacts on the 8$\times$ downsampling data, especially in dark and hair regions. Our fine-tuned model is able to remove such artifacts.

\begin{table}[t!]
  \centering
  \resizebox{\columnwidth}{!}{
      \begin{tabular}{l|c c c c|c}
        \toprule
  & PSNR$\uparrow$ & SSIM$\uparrow$ & LPIPS$\downarrow$ & FID$\downarrow$ & \multicolumn{1}{c}{\begin{tabular}[c]{@{}c@{}}Diffusion at \\ inference time?\end{tabular}} \\
\midrule
DifFace~\cite{yue2022difface}                   & 20.02 & 0.5225 & 0.6077 & 110.36 & \cmark \\ 
 DiffBIR~\cite{lin2023diffbir}                 & 20.28 & 0.4959 & 0.6605 & 126.44 & \cmark \\ 
PG-Diff~\cite{yang2023pgdiff}                  & 19.50 & 0.5339 & 0.5484 & 119.82 & \cmark\\ 
 DR2~\cite{wang2023dr2}                    & 20.34 & 0.5658 & 0.4815 & 54.94  & \cmark \\ 
 Our pseudo targets & \textbf{21.66} & \textbf{0.6044} & \textbf{0.4706} & \textbf{46.29} & \cmark \\ 
\midrule
\midrule
 GFPGAN~\cite{gfpgan}                    & 21.09 & 0.5283 & 0.5298 &  82.90 & \xmark \\ 
 VQFR~\cite{vqfr}                      & 19.38 & 0.4540 & 0.5567 & 123.12  &  \xmark \\ 
 CodeFormer~\cite{codeformer}             & 21.40 & 0.5412 & 0.4679 & 62.04   & \xmark \\ 
 CodeFormer + Ours  & \textbf{21.83} & \textbf{0.5681} & \textbf{0.4440} & \textbf{44.19} & \xmark \\ 
        \bottomrule
      \end{tabular}
      }
  \vspace{-5pt}
  \caption{\textbf{Results on data at 8$\times$ downsampling test set.} Top rows: diffusion-dependent models at test time. Bottom rows: diffusion-free models at test time.}
  \label{tab:synthetic_results}
  \vspace{-1.0em}
\end{table}

We show quantitative comparison for the two models in~\cref{tab:finetuning_results}. For SwinIR, the pre-trained model receives consistent improvements after fine-tuning. The perceptual improvements are  reflected in terms of the large boost in FID and LPIPS scores. For CodeFormer, we found it to be more resistant to smaller degradations (4$\times$ \textit{moderate}), and hence the fine-tuning mostly makes the images more realistic by removing the remaining artifacts (as observed in~\cref{fig:finetuning_synthetic} top two rows). On inputs with larger degradations (4$\times$ \textit{severe} and 8$\times$), our approach consistently improves the pre-trained model.

\begin{table}[t!]
  \centering
  \resizebox{\columnwidth}{!}{
      \begin{tabular}{l|c c c|c}
        \toprule
& MANIQA$\uparrow$ & MUSIQ$\uparrow$ & FID$\downarrow$ & \multicolumn{1}{c}{\begin{tabular}[c]{@{}c@{}}Diffusion at \\ inference time?\end{tabular}}\\
\midrule
DifFace~\cite{yue2022difface}                   & 0.5252 & 55.10 & 89.19  & \cmark\\ 
 DiffBIR~\cite{lin2023diffbir}                 & 0.5994 & 62.42 & 92.33 & \cmark \\ 
PG-Diff~\cite{yang2023pgdiff}                  & 0.5643 & 61.69 & 97.82  & \cmark \\ 
 DR2~\cite{wang2023dr2}                    & \textbf{0.6007} & \textbf{68.05} & 90.45 & \cmark \\ 
 Our pseudo targets                                 & 0.5772 & 62.56 & \textbf{80.60} & \cmark \\ 
\midrule
\midrule
 GFPGAN~\cite{gfpgan}                        & 0.5864 & 67.14 & 87.71  & \xmark \\ 
 VQFR~\cite{vqfr}                          & 0.5929 & 69.19 & 91.76 & \xmark \\ 
 CodeFormer~\cite{codeformer}                 & 0.6082 & 66.46 & 87.60  & \xmark \\ 
 CodeFormer + Ours & \textbf{0.6343} & \textbf{73.02} & \textbf{84.65}  & \xmark \\ 
        \bottomrule
      \end{tabular}
      }
  \vspace{-5pt}
  \caption{\textbf{Results on Wider-Test-200 set.} Top rows: diffusion-dependent models at test time. Bottom rows: diffusion-free models at test time.}
  \label{tab:wider_results}
  \vspace{-1.0 em}
\end{table}

\noindent \textbf{Comparison with state-of-the-art models.}
We benchmark our best fine-tuned model (fine-tuned CodeFormer) against other blind face restoration baselines in~\cref{tab:synthetic_results}. Our approach (bottom row) outperforms all baseline methods. In top rows we provide results of the zero-shot diffusion-based baselines that use a diffusion model at test time and compare them with our pseudo targets. Our targets improve on all metrics compared to these models, showing our approach can better preserve the content in the input image (higher PSNR and SSIM) while being more realistic according to LPIPS and FID scores.

The bottom rows in~\cref{tab:synthetic_results} show test-time non-diffusion based models, compared to which we improve consistently on all metrics. This group contains closer baseline models to ours, as none of them require running a diffusion model at test time, and hence are much more efficient. An interesting observation is that on some metrics our fine-tuned results are better than our pseudo targets. We believe a mixture of inductive bias from the pre-trained model together with distilled information from the diffusion model is injected into the parameters of the restoration model. In addition, directly learning from samples of the target (fine-tuning) dataset helps better generalize to target degradations. As shown in the first ablation study~(\cref{sec:ablation}), using more samples of the target dataset leads to better fine-tuned models. 

\noindent \textbf{Evaluation on real-world dataset.} In~\cref{tab:wider_results}, 
we compare our fine-tuned model with baselines on real-world degradations (Wider-Test-200). We improve on all non-reference based metrics compared to all baselines. We provide qualitative results on Wider-Test-200 in the supplementary material. One observation is that DR2 obtains higher MANIQA and MUSIQ while our approach gets higher FID. We believe that it is due to compromising fidelity for quality, as observed in
examples in the visual results of Wider-Test-200 in the supplementary material. Our fine-tuned model, however, improves on all metrics on this real-world dataset.

\subsection{Ablation Study}
\label{sec:ablation}

\begin{table}[t!]
  \centering
  \resizebox{\columnwidth}{!}{
      \begin{tabular}{c |c c c c | c c }
        \toprule
 Number of images & PSNR$\uparrow$ & SSIM$\uparrow$ & LPIPS$\downarrow$ & FID$\downarrow$ & MANIQA$\uparrow$ & MUSIQ$\uparrow$\\
\midrule
Pre-trained & 23.57 & 0.6391 & 0.4851 & 74.12 & 0.4623 & 64.38 \\
\midrule
\midrule
20         & 22.98 & 0.6321 & 0.4607 & 77.14 & 0.5239 & 59.05 \\
100        & 22.82 & 0.6218 & 0.4516 & 62.60 & 0.5748 & 68.56 \\
500        & 22.84 & 0.6216 & 0.4324 & 53.39 & 0.5932 & 72.17 \\
1000       & 23.10 & 0.6200 & 0.4286 & 51.11 & 0.5918 & 72.80 \\
2500       & 24.75 & 0.6676 & 0.3853 & 41.42 & 0.6023 & 73.36 \\
        \bottomrule
      \end{tabular}
    }
  \caption{\textbf{Effects of varying fine-tuning dataset size for SwinIR.} 
  }
  \label{tab:num_imgs_swinir}
  \vspace{-1.0 em}
\end{table}
\noindent \textbf{Effects of fine-tuning dataset size.} We investigate the effects of varying fine-tuning dataset size. In~\cref{tab:num_imgs_swinir}, we compare the results of the fine-tuning pre-trained SwinIR fine-tuning datasets of different sizes (i.e. different number of low-quality and pseudo target pairs). We starts to obtain consistent improvements even with only 100 fine-tuning images. The fine-tuned SwinIR becomes worse with 20 fine-tuning images. We believe that the SwinIR has over-fitted to this extremely small fine-tuning dataset in this case, which causes slight performance degradation. We also provide the same ablation on fine-tuned CodeFormer in~\cref{tab:num_imgs_codeformer}. 

\begin{figure}[t!]
\begin{center}
\includegraphics[width=1.0\linewidth]{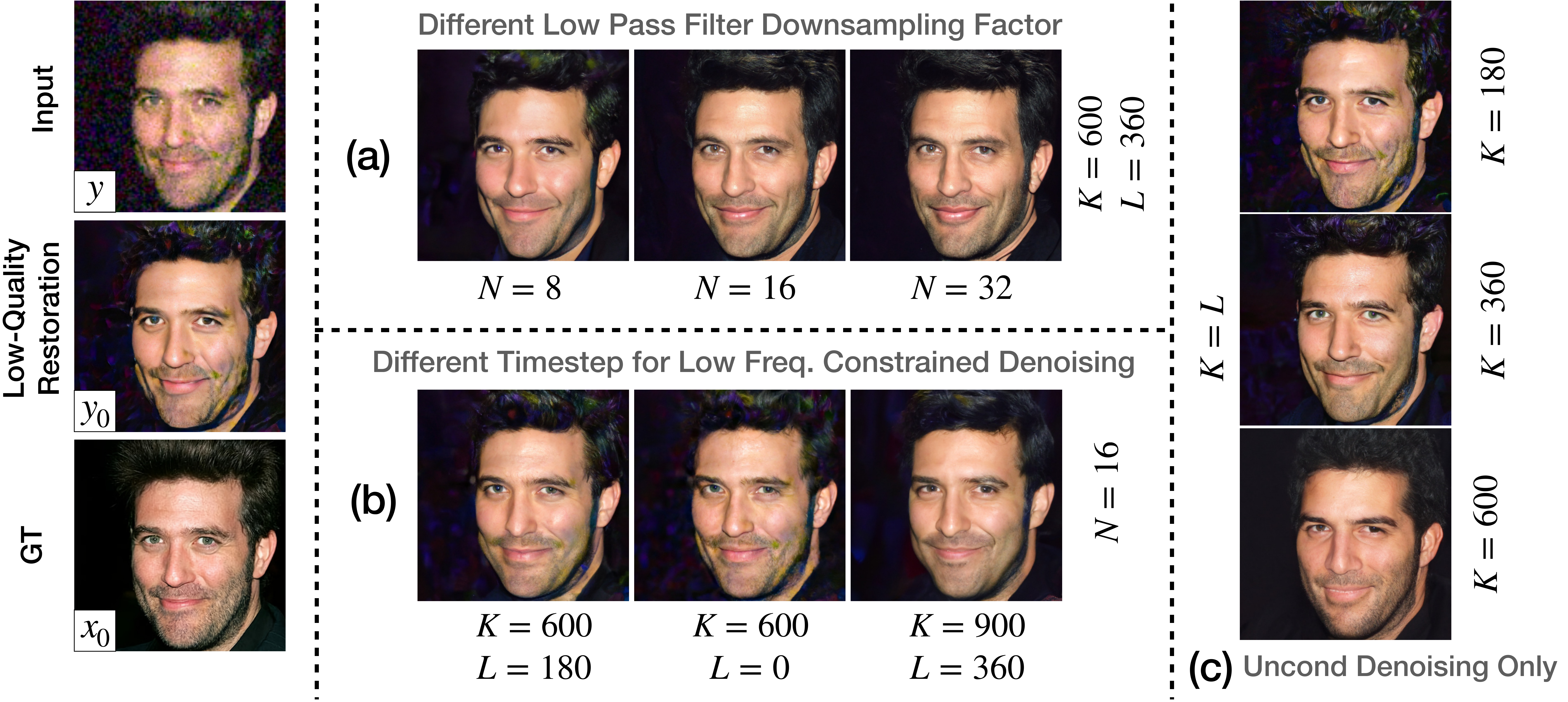}
\end{center}
\vspace{-10pt}
\caption{\textbf{Effects of low pass downsampling factor and timestep choices on pseudo target.} In (a) we show the effects of adjusting the low pass filter downsampling factor ($N$); In (b) we show the effects of different timestep windows for low frequency constrained denoising ($K$ and $L$); In (c) we show the pseudo targets if only unconditional denoising is applied (\textbf{zoom in for details}). A large version of this figure is included in the supplementary material for better visualization of the details.} 
\label{fig:targets_ablation}
\vspace{-1.0em}
\end{figure}
\noindent \textbf{Effects of timesteps and low pass filter choices.}
In the pseduo target generation stage, our approach involves applying a forward diffusion process on low-quality restoration to timestep $t=K$, and then utilizing low frequency constrained denoising process from timestep $t=K$ down to $t=L$, followed by the standard unconditional denoising for the rest of the timesteps. Having downsampling factor of $N=8$ for the low pass filter gives better fidelity to the inputs, but with artifacts carried over, while larger $N$ produces targets with higher perceptual quality. Overall, $N=16$ gives the best balance between fidelity and quality (\cref{fig:targets_ablation} (a)). As shown in~\cref{fig:targets_ablation} (b), applying low frequency constraint down to $L=180$ and to $L=0$ would also produce lower quality images due to artifacts in the low frequency content at small timesteps, while our approach is robust to the starting timestep ($K=900$). We also show the results of unconditional denoising for all timesteps in (c), where small timesteps of ($K=180$) and ($K=360$) produces artifacts, and a large timestep of ($K=600$) distorts the image content due to the lack of regularization during denoising.
\section{Conclusion}
\label{sec:conclusion}

In this paper, we propose an unsupervised approach for blind face restoration that addresses the problem of pre-trained restoration models failing on out-of-distribution degradations. Our method requires neither paired ground-truth high-quality images nor knowledge of the inputs' degradation process. Our two-stage pipeline starts by generating pseudo targets using a pre-trained diffusion model with a combination of low frequency constrained denoising and unconditional denoising. We then fine-tune the pre-trained models with pairs of low-quality inputs and pseudo targets. Our approach can consistently improve a pre-trained model's performance on out-of-distribution degradations, producing realistic restoration with satisfactory balance of fidelity and perceptual quality. Our best fine-tuned model achieves state-of-the-art blind face restoration on both synthetic and real-world datasets. 

\noindent \textbf{Acknowledgements.} This work was done at Samsung AI Center Toronto. It was funded by Mitacs and Samsung Research, Samsung Electronics Co., Ltd.

\appendix
\label{sec:appendix}

\section{Summary}
\label{sec:appendix_summary}

In~\cref{sec:synthetic_data_appendix}, we provide more details of our synthetic datasets.~\cref{sec:implementation_details_appendix} contains additional implementation details of pre-trained models and fine-tuning. We present additional qualitative and quantitative results along with more ablation studies in~\cref{sec:additional_results}. We provide more discussions on other relevant components of our approach in~\cref{sec:additionl_analysis}. Finally,~\cref{sec:limitations} and~\cref{sec:negative_effects} discuss the limitations and the potential negative impact of our method. 

\section{Details on Synthetic Datasets}
\label{sec:synthetic_data_appendix}

\subsection{Pre-training Dataset}

To perform the pre-training, we adopt the low-quality data synthesis pipeline from other literature~\cite{gfpgan,li2018learning,li2020blind} to create input-GT training data pairs: 
\begin{equation}
    \mathcal{I}_{LQ} = \Bigl\{\bigl[(\mathcal{I}_{HQ} * \mathbf{k}_{\sigma})_{\downarrow_{r}} + \mathbf{n}_{\delta}\bigr]_{JPEG_{q}}\Bigr\}_{\uparrow_{r}},
    \label{Eq:common_degradation_appendix}
\end{equation}
where a high-quality image $\mathcal{I}_{HQ}$ is first convolved with a Gaussian blur kernel $\mathbf{k}_{\sigma}$ of kernel size $\sigma$ and downsampled by a factor of $r$. Gaussian noise with standard deviation of $\delta$ is added to the downsampled image. Then JPEG compression of quality factor $q$ and upsampling by a factor of $r$ are applied to synthesize the low-quality counterpart $I_{LQ}$. For pre-training the SwinIR~\cite{liang2021swinir} model, we use randomly sampled $\sigma$, $r$, $\delta$, and $q$ from $[0.1, 15]$, $[0.8, 32]$, $[0, 20]$, and $[30, 100]$, respectively. We use the pre-trained CodeFormer~\cite{codeformer} from the authors. This model is pre-trained with randomly sampled $\sigma$, $r$, $\delta$, and $q$ from $[1, 15]$, $[1, 30]$, $[0, 20]$, and $[30, 90]$.

\subsection{Our Synthetic Dataset}
As mobile phones become more accessible, significant number of images are captured with phone cameras nowadays. We can simulate more realistic low-quality images by taking the camera ISP and RAW noise models into consideration. Specifically, we generate the low-quality images from high-quality ones as:
\begin{equation}
    \mathcal{I}_{LQ} = \Bigl\{ISP\bigl[ISP^{-1}((\mathcal{I}_{HQ})_{\downarrow_{r}}) + \mathbf{n}_{c}\bigr]\Bigr\}_{\uparrow_{r}},
    \label{Eq:realistic_degradation_appendix}
\end{equation}
where ($ISP^{-1}$) is an image unprocessing pipeline~\cite{seo2023graphics2raw} that converts a sRGB image to RAW, ($ISP$) is the reverse procedure of image unprocessing to render the sRGB image from RAW image~\cite{seo2023graphics2raw}, $r$ is the downsampling and upsampling factor, and $n_c$ is the simulated RAW camera noise. For the simulated noise, we apply the commonly used Heteroscedastic Gaussian models~\cite{foi2008practical,liu2014practical,makitalo2012optimal} to generate realistic camera noise. We construct our synthetic datasets at \textit{moderate} and \textit{severe} noise levels, at ISO levels of ($\sim1600$) and ($\sim3200$), respectively. For each noise level, we also generate two separate datasets at 4$\times$ downsampling and 8$\times$ downsampling, which gives us four datasets in total. For each dataset, we use the same 3,000 high-quality images from the CelebA-HQ dataset~\cite{celebahq}. We take 2,500 of them for generating pesudo targets and fine-tuning the pre-trained models, and the rest of the 500 images for evaluation. We do not use the GT images during pseudo target generation and fine-tuning. For fine-tuning, we start with the same pre-trained checkpoint for all four dataset setups.

\begin{table*}[ht]
  \centering
  \addtolength{\tabcolsep}{3pt}
      \begin{tabular}{c c c|c c c c | c c }
        \toprule
Encoder & Trans. & CFT & PSNR$\uparrow$ & SSIM$\uparrow$ & LPIPS$\downarrow$ & FID$\downarrow$ & MANIQA$\uparrow$ & MUSIQ$\uparrow$\\
\midrule
\cmark                   &  &  & 22.38 & 0.5816 & 0.4391 & \textbf{40.59} & 0.6568 & 75.60 \\ 
\cmark                 & \cmark & & 22.85 & 0.6036 & 0.4258 & 41.21 & \textbf{0.6581} & \textbf{75.76} \\ 
\cmark                  & \cmark & \cmark & \textbf{23.01} & \textbf{0.6556} & \textbf{0.4225} & 47.67 & 0.6295 & 71.13 \\ 
        \bottomrule
      \end{tabular}
  \vspace{-5pt}
  \caption{\textbf{Different fine-tuning setups for CodeFormer.} We compare the three setups of fine-tuning CodeFormer on 4$\times$ downsampling data at \textit{severe} noise level.}
  \label{tab:codeformer_finetune_setup}
  \vspace{-1.0 em}
\end{table*}

\section{Implementation Details}
\label{sec:implementation_details_appendix}

\subsection{Pre-trained Models}
\label{sec:pretrained_appendix}

In our work, both SwinIR~\cite{liang2021swinir} and CodeFormer~\cite{codeformer} are pre-trained on the entire FFHQ dataset~\cite{ffhq}, which contains 70,000 high-quality face images with resolution of 512$\times$512. The training data pairs are generated by following the degradation pipeline described in~\cref{Eq:common_degradation_appendix}.

We train the SwinIR from scratch using the AdamW optimizer~\cite{loshchilov2017decoupled} with initial learning rate of 1$e-4$ and update the learning rate following cosine annealing~\cite{loshchilov2016sgdr}. We use batch size of 16 and train the model for 800,000 iterations in total with image-level $L1$ loss, perceptual (LPIPS) loss~\cite{lpips}, and adversarial (GAN) loss~\cite{goodfellow2014generative}. For losses, we use the same set of losses and the weights as the fine-tuning setup described in the main paper.

For CodeFormer, we directly adopt the pre-trained checkpoint from the CodeFormer authors~\cite{codeformer}. It is pre-trained following their 3-stage training scheme. Please refer to their paper for more details on CodeFormer pre-training.

For the unconditional face diffusion model used in our pipeline, we follow the model architecture from~\cite{guided} and the adopt the model weights from~\cite{yue2022difface}. This model is also trained on FFHQ dataset~\cite{ffhq}. The total number of timesteps is $T=1000$ and we follow techniques from~\cite{nichol2021improved} to accelerate the denoising process by reducing the total number discrete timesteps down to 250.

\subsection{Fine-tuning}

For fine-tuning the pre-trained SwinIR model, we use the same optimizer, initial learning rate, and batch size as in pre-training (\cref{sec:pretrained_appendix}). We fine-tune the SwinIR model with input and pseudo target pairs for 20,000 iterations. The setup for the loss functions has been described in the main paper.

For CodeFormer, we only fine-tune its encoder and transformer module while keeping the codebook, decoder, and the controllable feature transformation module (CFT) fixed. We use the code-level loss $\mathcal{L}_{code}^{feat}$ and the cross-entropy loss on the code prediction $\mathcal{L}_{code}^{token}$ to fine-tune the encoder and transformer module (same set of losses used in stage-II training of CodeFormer):
\begin{equation}
    \mathcal{L} = \mathcal{L}_{code}^{feat} + \lambda_{token}\mathcal{L}_{code}^{token},
    \label{Eq:codeformer_finetune_loss_definitions}
\end{equation}
where $\lambda_{token}$ is the weight for the cross-entropy loss on the code prediction.

Note that no image-level loss is used since we found empirically that only fine-tuning the encoder and transformer module provides the best performance, and there is no need for image-level losses when only fine-tuning these two parts of the model. We compare the results of different fine-tuning setups in~\cref{tab:codeformer_finetune_setup}. Fine-tuning the encoder and the transformer module provides the best balance between fidelity and quality among all the setups. Fine-tuning all three components obtains better PSNR, SSIM, and LPIPS. However, there is a compromise in the image quality as reflected in FID, MANIQA, and MUSIQ metrics.

\begin{figure*}[t!]
\begin{center}
\includegraphics[width=1.0\linewidth]{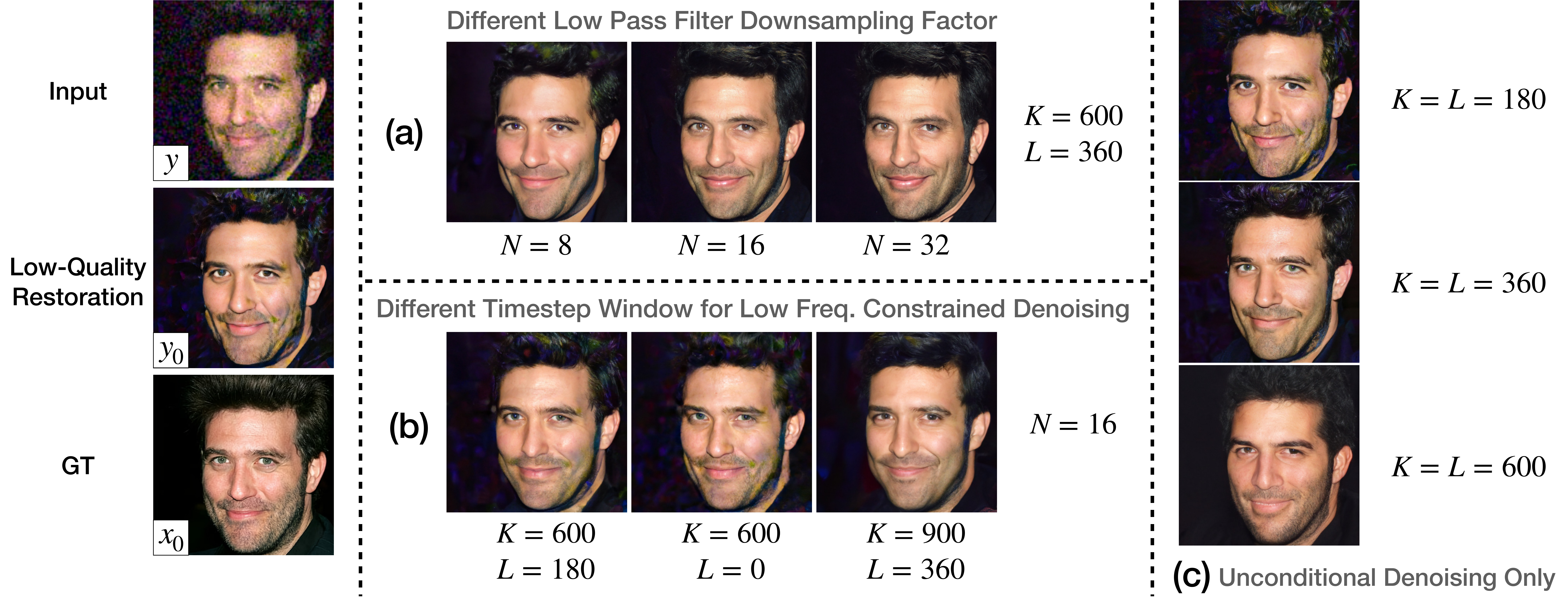}
\end{center}
\vspace{-10pt}
\caption{\textbf{Effects of low pass downsampling factor and timestep choices on SwinIR pseudo target.} In (a) we show the effects of adjusting the low pass filter downsampling factor ($N$); In (b) we show the effects of different timestep windows for low frequency constrained denoising ($K$ and $L$); In (c) we show the pseudo targets if only unconditional denoising is applied (\textbf{zoom in for details}).} 
\label{fig:swinir_targets_ablation}
\end{figure*}

\begin{figure*}[t!]
\begin{center}
\includegraphics[width=1.0\linewidth]{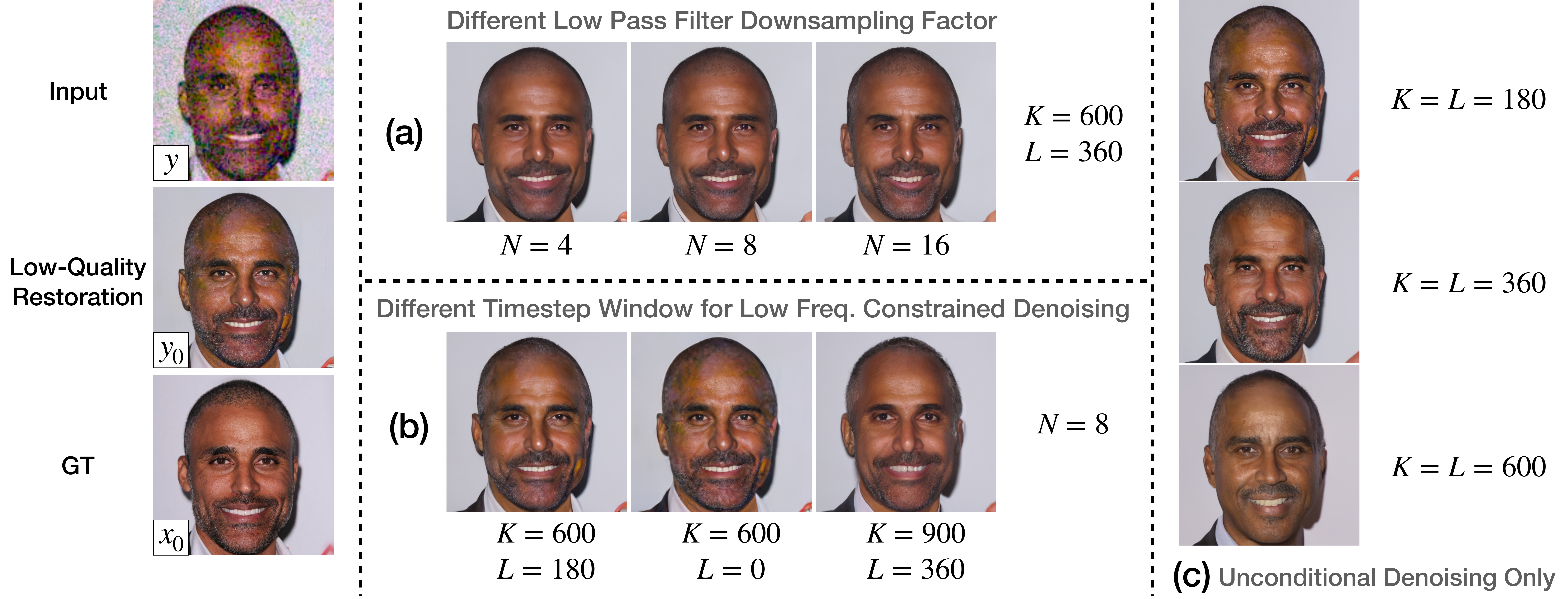}
\end{center}
\vspace{-10pt}
\caption{\textbf{Effects of low pass downsampling factor and timestep choices on CodeFormer pseudo target.} In (a) we show the effects of adjusting the low pass filter downsampling factor ($N$); In (b) we show the effects of different timestep windows for low frequency constrained denoising ($K$ and $L$); In (c) we show the pseudo targets if only unconditional denoising is applied (\textbf{zoom in for details}).}
\label{fig:codeformer_target_ablation_appendix}
\end{figure*}

\section{Additional Results}
\label{sec:additional_results}

\subsection{Qualitative Results}
\label{sec:qualitative_results_appendix}

\begin{figure*}[ht]
\begin{center}
\includegraphics[width=1.0\linewidth]{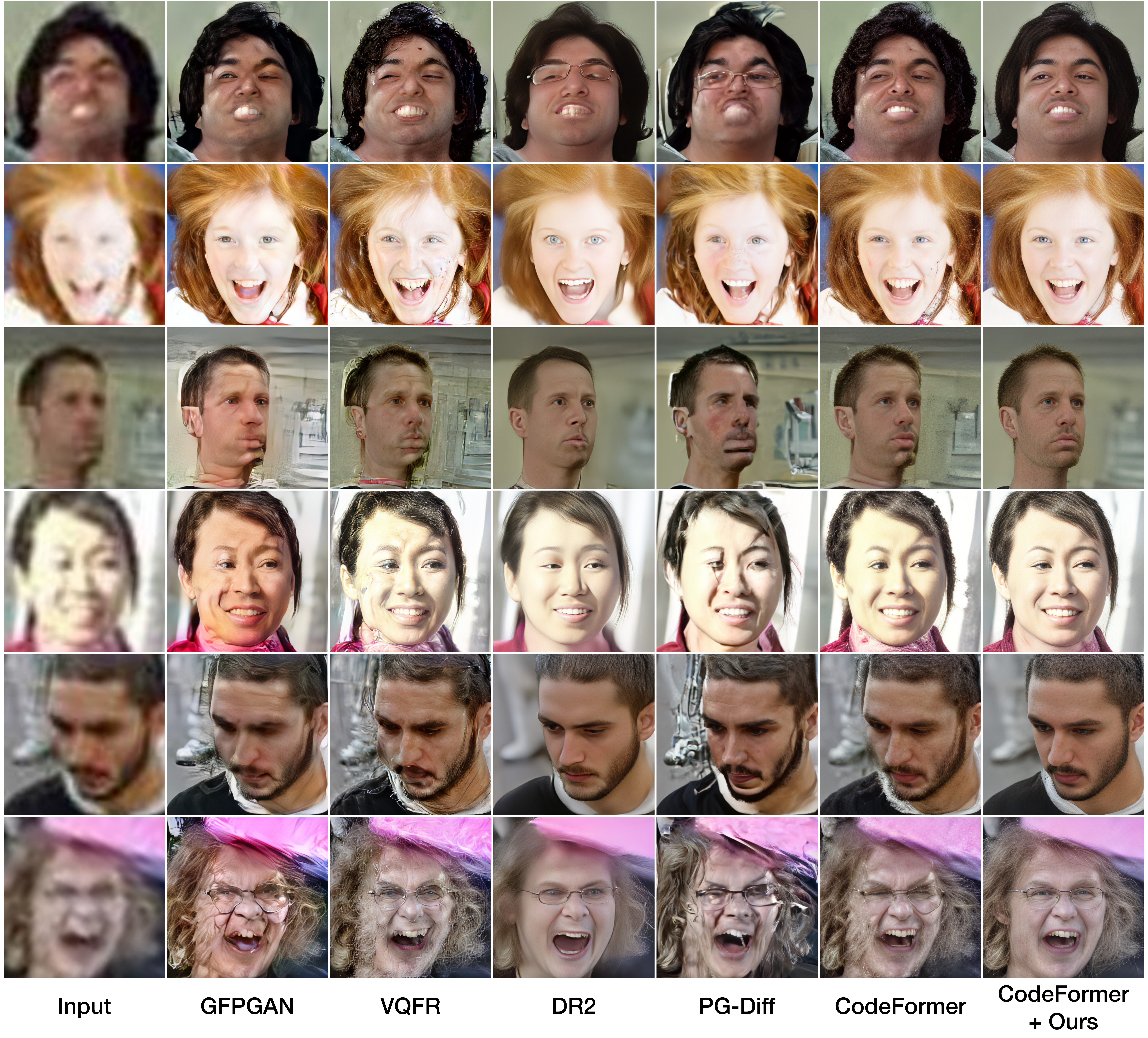}
\end{center}
\vspace{-10pt}
\caption{Qualitative comparison on testing samples from our Wider-Test-200 (\textbf{zoom in for details}).}
\label{fig:wider_appendix}
\vspace{-1.0em}
\end{figure*}

\begin{figure*}[ht]
\begin{center}
\includegraphics[width=1.0\linewidth]{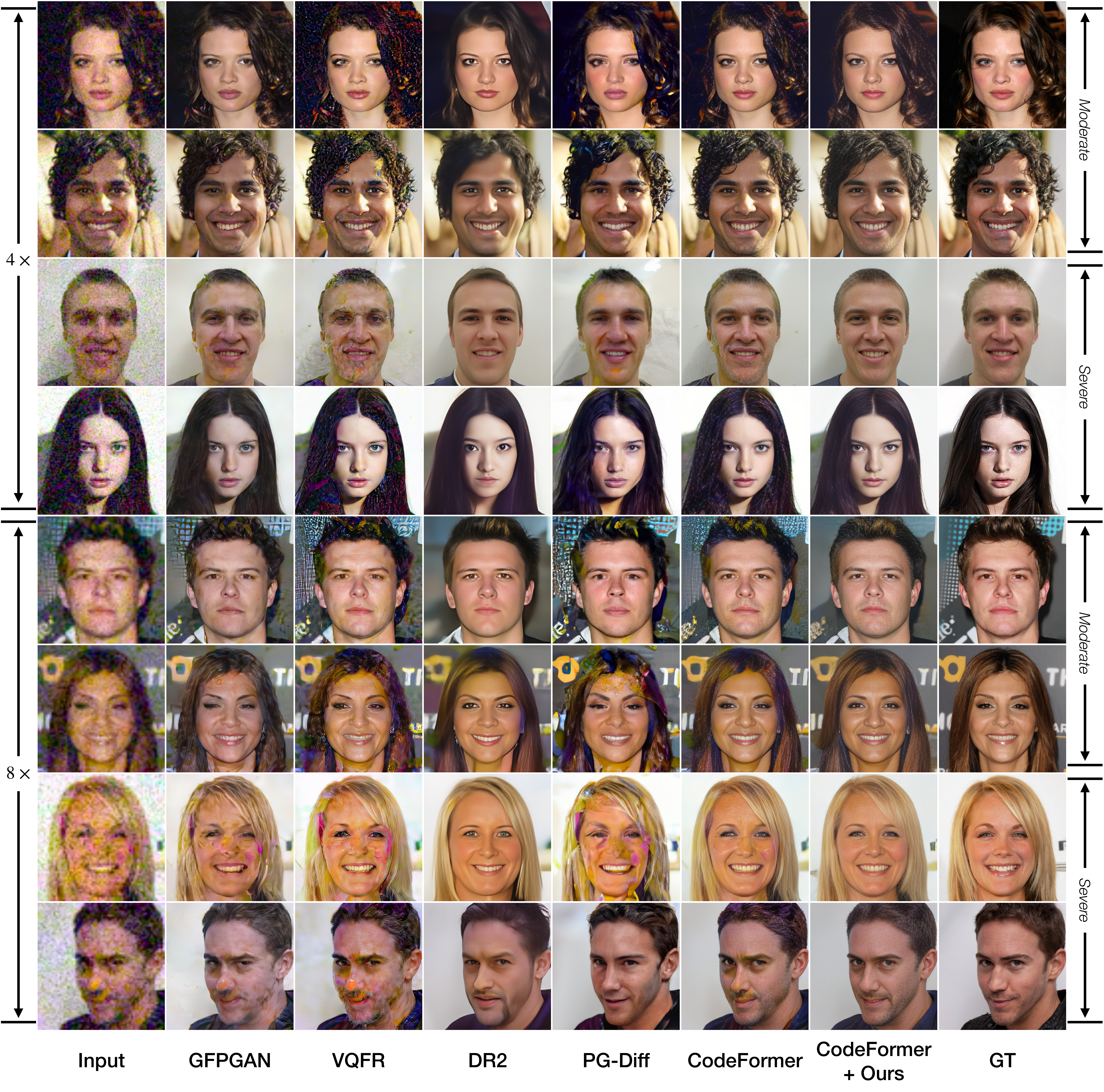}
\end{center}
\vspace{-10pt}
\caption{Additional qualitative comparison with other baselines at different degradation levels (\textbf{zoom in for details}).}
\label{fig:result_synthetic_appendix}
\vspace{-1.0em}
\end{figure*}

We show additional qualitative results on our synthetic datasets in~\cref{fig:finetuning_appendix,fig:result_synthetic_appendix}. We also provide qualitative results on our Wider-Test-200 set in~\cref{fig:wider_appendix}.

\subsection{Quantitative Results}
\label{sec:quantitative_results_appendix}

We provide detailed results for both SwinIR and CodeFormer on each synthetic dataset setup. The results of the pre-trained, fine-tuned SwinIR along with the SwinIR-based targets are included in~\cref{tab:swinir_appendix}. For CodeFormer, we provide the results in a similar manner as in the main paper, but for each noise level separately. That is, we show the CodeFormer results in~\cref{tab:moderate_codeformer_appendix,tab:severe_codeformer_appendix} for \textit{moderate} and \textit{severe} noise levels, respectively. Note that for all the models, we adopt the scripts and checkpoints from their authors, and run the models with their suggested parameters and setups for blind face restoration. The pre-trained CodeFormer achieves the best results on 4$\times$ \textit{moderate} noise levels because it is robust enough to handle such inputs, however it cannot completely get rid of all artifacts (See~\cref{fig:finetuning_appendix}). Fine-tuning this model effectively removes the remaining artifacts, but would slightly degrade its quantitative performance due to small content distortion in pseudo targets. However, on more severe degradations where the pre-trained model's performance degrades, our approach consistently improves upon the pre-trained model. In addition, we provide the results of fine-tuning the pre-trained model using GT clean images as pseudo targets (i.e. supervised fine-tuning) for both model architectures in~\cref{tab:swinir_gt_targets_appendix,tab:codeformer_gt_targets_appendix}. They serve as the performance upper bound for our problem setup.

\subsubsection{Mild degradations}
For completeness, we follow the same approach as \textit{medium} and \textit{severe} degradations to construct a fine-tuning dataset with weaker degradation using ISO level of ($\sim800$). We refer it as \textit{mild} degradation and report the results in~\cref{tab:swinir_mild_degradation}. The results show consistent improvement of our fine-tuned model compared to the pre-trained ones.

\subsubsection{Comparison on face recognition metrics}
We report two face recognition metrics: Deg.~\cite{gfpgan_comp} and LMD~\cite{vqfr_comp} in~\cref{tab:facial_recognition_metrics}. Our pseudo targets obtain the best LMD and the second best Deg. Among the diffusion-free models, our model obtains the best results on PSNR, SSIM, LPIPS, and FID, while obtaining lower performance on LMD and Deg. We observe that the models have to make a compromise between fidelity and quality. While the facial lankmark is deteriorated slightly, the reference-based metrics of PSNR, SSIM, and LPIPS still show superior performance of our proposed approach.

\subsubsection{Runtime and memory efficiency}
 In~\cref{tab:runtime_memory_comparison}, we compare the memory usage and inference time of our model with others on a Nvidia RTX 3090 GPU. For diffusion-based models, even if we reduce the number of time-steps to only one step, our method is more efficient in terms of both run-time and memory. This is still regardless of the drop in accuracy of diffusion-based models that one-step diffusion would incur, as shown in~\cite{meng2023distillation_comp, song2023consistency_comp}.

\subsection{Additional Ablation Studies}
\label{sec:additional_ablation}

\subsubsection{Effects of timesteps and low pass filter choices} 

The choices for timesteps of when to start and stop the low frequency content constraint and the low pass filter's downsampling factor have significant impact on the quality and fidelity of the pseudo targets. In~\cref{fig:swinir_targets_ablation,fig:codeformer_target_ablation_appendix}, we show the effects of the varying these parameters on SwinIR and CodeFormer pseudo targets. We provide quantitative ablation studies on the pseudo targets for both SwinIR~\cite{liang2021swinir} and CodeFormer~\cite{codeformer} in~\cref{tab:swinir_targets_ablation,tab:codeformer_targets_ablation}, as well as the results of fine-tuning using the corresponding targets in~\cref{tab:swinir_targets_ablation_finetuning,tab:codeformer_targets_ablation_finetuning}. Our selection of the parameters provide the best balance between perceptual quality and fidelity.

\subsubsection{Loss functions setup for fine-tuning}

\begin{table*}[ht]
  \centering
  \addtolength{\tabcolsep}{1pt}
      \begin{tabular}{c c c|c c c c | c c }
        \toprule
$\mathcal{L}_1$ & $\mathcal{L}_{LPIPS}$ & $\mathcal{L}_{GAN}$ & PSNR$\uparrow$ & SSIM$\uparrow$ & LPIPS$\downarrow$ & FID$\downarrow$ & MANIQA$\uparrow$ & MUSIQ$\uparrow$\\
\midrule
\cmark                   &  &  & 25.06 & \textbf{0.7207} & 0.4587 & 92.04 & 0.3151 & 41.72 \\ 
\cmark                 & \cmark & & \textbf{25.30} & 0.7106 & 0.3908 & 47.94 & 0.5545 & 59.30 \\ 
\cmark                  &  & \cmark & 24.52 & 0.6587& 0.4226 & 50.70 & 0.5869 & 72.88 \\ 
\cmark                  & \cmark & \cmark & 24.75 & 0.6676 & \textbf{0.3853} & \textbf{41.42} & \textbf{0.6023} & \textbf{73.36} \\ 
        \bottomrule
      \end{tabular}
  \caption{\textbf{Different setups of fine-tuning losses for SwinIR.} We compare different loss functions setup of fine-tuning SwinIR on 4$\times$ downsampling data at \textit{moderate} noise level.}
  \label{tab:loss_ablation}
\end{table*}

\begin{table*}[ht]
  \centering
  \addtolength{\tabcolsep}{3.5pt}
      \begin{tabular}{c c c|c c c c | c c }
        \toprule
$\mathcal{L}_1$ & $\mathcal{L}_{LPIPS}$ & $\mathcal{L}_{GAN}$ & PSNR$\uparrow$ & SSIM$\uparrow$ & LPIPS$\downarrow$ & FID$\downarrow$ & MANIQA$\uparrow$ & MUSIQ$\uparrow$\\
\midrule
1.0 & 1.0  & 1.0  & 24.18 & 0.6656 & 0.3913 & 42.57 & 0.6022 & 72.99 \\
1.0 & 0.1  & 1.0  & 23.83 & 0.6362 & 0.3881 & \textcolor{blue}{41.80} & \textcolor{blue}{0.6107} & 73.47 \\
1.0 & 1.0  & 0.1  & 24.89 & 0.6899 & 0.3873 & 46.37 & 0.6000 & 69.83 \\
\textbf{1.0} & \textbf{0.1}  & \textbf{0.1}  & 24.75 & 0.6676 & \textcolor{blue}{0.3853} & \textcolor{red}{41.42} & 0.6023 & 73.36 \\
1.0 & 0.01 & 1.0  & 23.16 & 0.6133 & 0.4034 & 46.48 & 0.5943 & \textcolor{blue}{73.82} \\
1.0 & 0.01 & 0.1  & 24.39 & 0.6531 & 0.3889 & 41.81 & \textcolor{red}{0.6247} & \textcolor{red}{74.77} \\
1.0 & 0.01 & 0.01 & 24.87 & 0.6801 & 0.3890 & 44.36 & 0.6020 & 71.10 \\
1.0 & 0.1  & 0.01 & \textcolor{blue}{25.16} & \textcolor{blue}{0.6943} & \textcolor{red}{0.3850} & 45.82 & 0.6002 & 69.54 \\
1.0 & 1    & 0.01 & \textcolor{red}{25.17} & \textcolor{red}{0.7041} & 0.3890 & 46.73 & 0.5863 & 65.11 \\
        \bottomrule
      \end{tabular}
  \vspace{5pt}
  \caption{\textbf{Ablation study on the weights of the fine-tuning losses for SwinIR.} We compare the results of fine-tuning pre-trained SwinIR with different loss weights on 4$\times$ downsampling data at  \textit{moderate} noise level. \textcolor{red}{Red} and \textcolor{blue}{blue} indicate the best and the second best results. \textbf{Bold} indicates our selection.}
  \label{tab:loss_weights_ablation_swinir}
\end{table*}

\begin{table*}[ht]
  \centering
  \addtolength{\tabcolsep}{3.5pt}
      \begin{tabular}{c c |c c c c | c c }
        \toprule
$\mathcal{L}_{code}^{feat}$ & $\mathcal{L}_{code}^{token}$ & PSNR$\uparrow$ & SSIM$\uparrow$ & LPIPS$\downarrow$ & FID$\downarrow$ & MANIQA$\uparrow$ & MUSIQ$\uparrow$\\
\midrule
1.0 & 1.0 & \textbf{22.31} & \textbf{0.5856} & \textbf{0.4282} & 42.16 & 0.6574 & \textbf{75.84} \\
\textbf{1.0} & \textbf{0.5} & 22.28 & 0.5848 & 0.4290 & \textbf{41.72} & \textbf{0.6580} & \textbf{75.84} \\
1.0 & 0.1 & 22.26 & 0.5833 & 0.4303 & 42.18 & 0.6578 & 75.81 \\
        \bottomrule
      \end{tabular}
  \vspace{5pt}
  \caption{\textbf{Ablation study on the weights of the fine-tuning losses for CodeFormer.} We compare the results of fine-tuning pre-trained CodeFormer with different weights on 4$\times$ downsampling data at  \textit{severe} noise level. \textbf{Bold} indicates our selection.}
  \label{tab:loss_weights_ablation_codeformer}
\end{table*}

The performance of our approach depends on the fine-tuning of the pre-trained model. In~\cref{tab:loss_ablation}, we provide the results of a fine-tuned SwinIR model with different combinations of the fine-tuning losses. We use the same set of input and pseudo target data pairs for all the setups in this table and fine-tune the model for the same number of iterations. Using $\mathcal{L}_{L1}$ and $\mathcal{L}_{LPIPS}$ together obtains better PSNR, however, it obtains worse results on perceptual metrics. When applying all the three losses, the model's results are in the best perceptual quality, achieving the best LPIPS and FID among all the setups. We provide the ablation on the fine-tuning loss functions for CodeFormer in the supplementary material.

\subsubsection{Weights of the loss functions} 

In~\cref{tab:loss_weights_ablation_swinir}, we compare the results of fine-tuning SwinIR with different weights for the perceptual loss ($\mathcal{L}_{LPIPS}$) and adversarial loss ($\mathcal{L}_{GAN}$). In our experiments, we set the weights of these two losses to be $0.1$, as this setup gives the best perceptual quality among all the combinations while maintaining good fidelity. We also conduct ablation study on the weight of the cross-entropy loss ($\mathcal{L}_{code}^{token}$) for CodeFormer fine-tuning. As shown in~\cref{tab:loss_weights_ablation_codeformer}, the results of the fine-tuned model do not vary much with different weights.

\begin{table*}[ht]
  \centering
  \addtolength{\tabcolsep}{3.5pt}
      \begin{tabular}{c |c c c c | c c }
        \toprule
 Number of images & PSNR$\uparrow$ & SSIM$\uparrow$ & LPIPS$\downarrow$ & FID$\downarrow$ & MANIQA$\uparrow$ & MUSIQ$\uparrow$\\
\midrule
Pre-trained & 22.90 & 0.5810 & 0.4420 & 53.02 & 0.6371 & 74.96 \\
\midrule
\midrule
20         & 23.42 & 0.6288 & 0.4276 & 46.08 & 0.6535 & 74.93 \\
100        & 23.28 & 0.6189 & 0.4259 & 44.25 & 0.6562 & 75.40 \\
500        & 22.99 & 0.6080 & 0.4262 & 42.34 & 0.6590 & 75.63 \\
1000       & 22.88 & 0.6038 & 0.4266 & 41.81 & 0.6590 & 75.76 \\
2500       & 22.85 & 0.6036 & 0.4258 & 41.21 & 0.6581 & 75.76 \\
        \bottomrule
      \end{tabular}
  \vspace{5pt}
  \caption{\textbf{Ablation Study on the fine-tuning dataset size on CodeFormer.} We compare the results of fine-tuning pre-trained CodeFormer with different numbers of images used in fine-tuning on 4$\times$ downsampling data at \textit{severe} noise level.}
  \label{tab:num_imgs_codeformer}
\end{table*}

\subsubsection{Number of images used in fine-tuning} 
We investigate the effects of the fine-tuning dataset size on both SwinIR and CodeFormer. In~\cref{tab:num_imgs_swinir,tab:num_imgs_codeformer}, we compare the results of the fine-tuning models with pre-trained models using different sizes of the fine-tuning dataset. We gain consistent improvements with our approach on CodeFormer even with 20 images, thanks to the discrete codebook that prevents potential over-fitting. 

\begin{figure*}[ht]
\begin{center}
\includegraphics[width=1.0\linewidth]{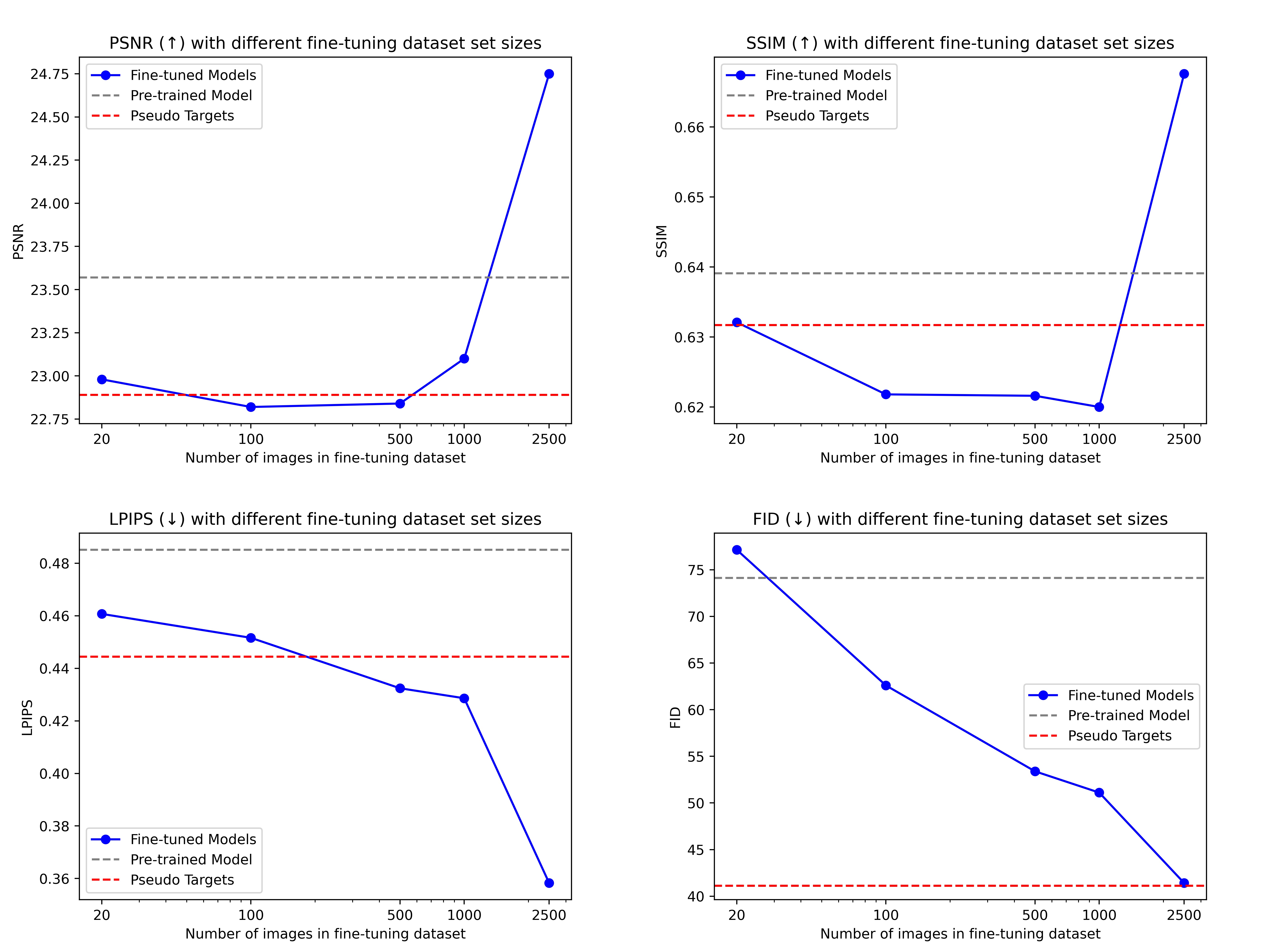}
\end{center}
\vspace{-10pt}
\caption{Fine-tuned SwinIR's performance when using different sizes of fine-tuning datasets on 4$\times$ downsampling data at \textit{moderate} noise level.}
\label{fig:num_images_ablation_graphs}
\vspace{-1.0em}
\end{figure*}

In~\cref{fig:num_images_ablation_graphs}, we evaluate the impact of gradually increasing the number of images used in fine-tuning on the fine-tuned restoration model performance. As can be observed, more images help improve the restoration model beyond the pseudo targets. In particular, training on such images adds a prior to the restoration model on how to handle the observed artifacts, allowing the model to learn on the ensemble of target degradations, which is not the case for pseudo-targets.

\section{More Analysis and Discussions}
\label{sec:additionl_analysis}

\subsection{Pseudo Target Generation vs. Other Diffusion-based Methods} 

Here we provide more detailed comparison between our pseudo target generation process and other relevant diffusion-based methods. Specifically, we compare the detailed procedure of our pseudo target generation with DifFace~\cite{yue2022difface}, ILVR~\cite{ilvr}, DDA~\cite{gao2023back}, PG-Diff~\cite{yang2023pgdiff}, DiffBIR~\cite{lin2023diffbir}, and DR2~\cite{wang2023dr2}. We summarize the detailed procedures in~\cref{alg:sampling_ours,alg:sampling_difface,alg:sampling_ilvr,alg:sampling_dda,alg:sampling_pgdiff,alg:sampling_diffbir,alg:sampling_dr2}, respectively. 

To highlight the main differences, DifFace~\cite{yue2022difface} adds Gaussian noise to the output of a preprocessing model~\cite{liang2021swinir}, and applies standard unconditional denoising to get the clean image. ILVR~\cite{ilvr} and DDA~\cite{gao2023back} apply similar low frequency content guidance, but for all the denoising timesteps. Note that despite ILVR targeting tasks of conditional image generation and image editing, and DDA targeting domain adaptation for image classification, their denoising diffusion process also utilizes low frequency content as guidance, which in theory could be applied to the task of blind image restoration. However, as they apply such guidance for all denoising timesteps, the results would be blurry and still contain artifacts due to inaccurate guidance from degraded input image.  PG-Diff~\cite{yang2023pgdiff} and DiffBIR~\cite{lin2023diffbir} use the output of a pre-processing model~\cite{wang2021realesrgan, liang2021swinir} on the input image to guide the denoising process. PG-Diff uses an unconditional face diffusion model~\cite{guided}, while DiffBIR uses a fine-tuned stable-diffusion model~\cite{rombach2022high}. DR2~\cite{wang2023dr2} stops the low frequency content constraining at a pre-defined timestep, performs a one-step project to timestep $t=0$, and relies on a pre-trained restoration model~\cite{vqfr} as a post-processing step to restore the high-frequency details. It also downsamples the input by a factor of 2 before applying a 256$\times$256 face diffusion model, and upsamples the resulting image from the diffusion model to restore the original resolution before applying the pre-trained post-processing model. This downsampling procedure helps reducing the amount of artifacts in the image before passing it to the diffusion model, at the expense losing finer details.

Since DR2~\cite{wang2023dr2} achieves comparable performance as our targets on Wider-Test-200 set (better on MANIQA and MUSIQ while worse on FID), we perform experiments of fine-tuning pre-trained models with DR2's results as pseudo targets. We compare the results of fine-tuning SwinIR and CodeFormer with our pseudo targets and DR2 outputs in~\cref{tab:finetuning_with_dr2_targets}. With our pseudo targets, both fine-tuned models can produce more realistic results. 

\begin{table}[t!]
  \centering
  \resizebox{\columnwidth}{!}{
      \begin{tabular}{c|c|c c c}
        \toprule
Model &  Targets & MANIQA$\uparrow$ & MUSIQ$\uparrow$ & FID$\downarrow$\\
    \midrule

    \multirow{2}{*}{Fine-tuned SwinIR} & DR2 & 0.5740 & 69.27 & 89.48 \\
     & Ours & \textbf{0.6093} & \textbf{74.15} & \textbf{88.21} \\
    \midrule
    \midrule
    \multirow{2}{*}{Fine-tuned CodeFormer} & DR2 & \textbf{0.6386} & 72.63 & 90.59\\
     & Ours & 0.6343 & \textbf{73.02} & \textbf{84.65} \\
        \bottomrule
      \end{tabular}
    }
  \vspace{-5pt}
  \caption{\textbf{Our pseudo targets vs. DR2~\cite{wang2023dr2} as targets.} We compare the results of fine-tuning models with our pseudo targets and with DR2 outputs as targets on the Wider-Test-200 set.}
  \label{tab:finetuning_with_dr2_targets}
\end{table}

\begin{figure*}[ht]
\begin{center}
\includegraphics[width=1.0\linewidth]{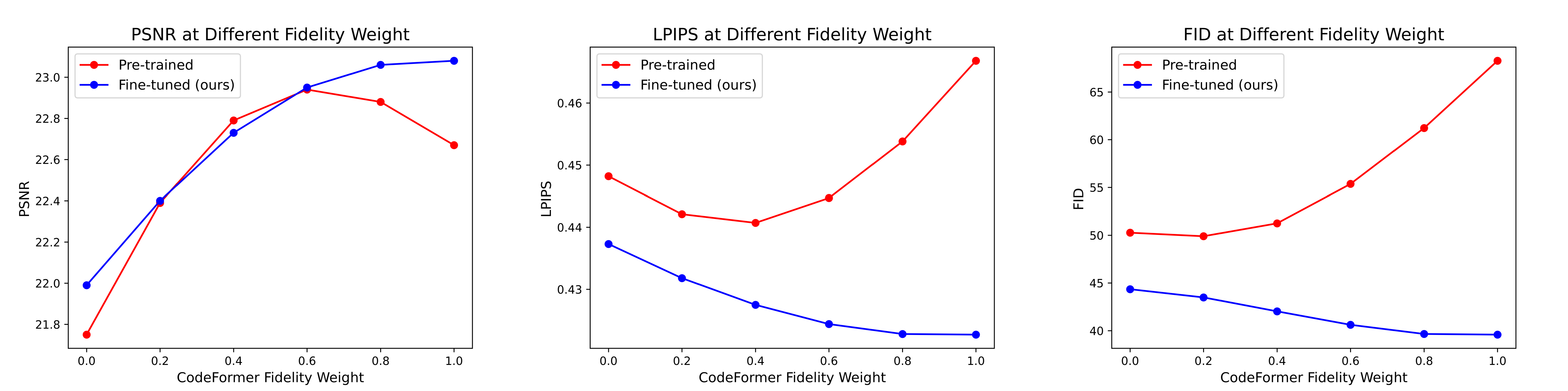}
\end{center}
\vspace{-10pt}
\caption{\textbf{CodeFormer results using different fidelity weights.} We plot the results of pre-trained and fine-tuned CodeFormer when using different fidelity weights at testing time on 4$\times$ downsampling data at \textit{severe} noise level.}
\label{fig:fidelity_curve_appendix}
\end{figure*}

\subsection{CodeFormer Fidelity Weight}

CodeFormer~\cite{codeformer} has a unique controllable features transformation (CFT) module that controls how much information from the encoder is fused into the decoder. It allows users to control the balance between the quality ($w=0$) and fidelity ($w=1$) of the restoration, by simplying modifying this fidelity weight $w \in [0, 1]$. We set the $w=0.5$ (default value suggested by CodeFormer) when we run the pre-trained and the fine-tuned CodeFormer in this paper for a fair comparison between the two. We also show the comparison of our fine-tuned CodeFormer with its pre-trained counterpart at different fidelity weights in~\cref{fig:fidelity_curve_appendix}. Our fine-tuned model is consistently better than the pre-trained model. In addition, the fine-tuned model achieves the best results at $w=1.0$. We believe that fine-tuning improves the quality of the extracted features from the encoder, thus more information from the encoder flowing into the decoder is helpful in terms of both restoration quality and fidelity.

\subsection{Pseudo Target Generation Without Pre-trained Restoration Model}

\begin{figure}[t!]
    \centering
    \includegraphics[width=\linewidth, trim={0cm 0.1cm 0cm 0.2cm},clip]{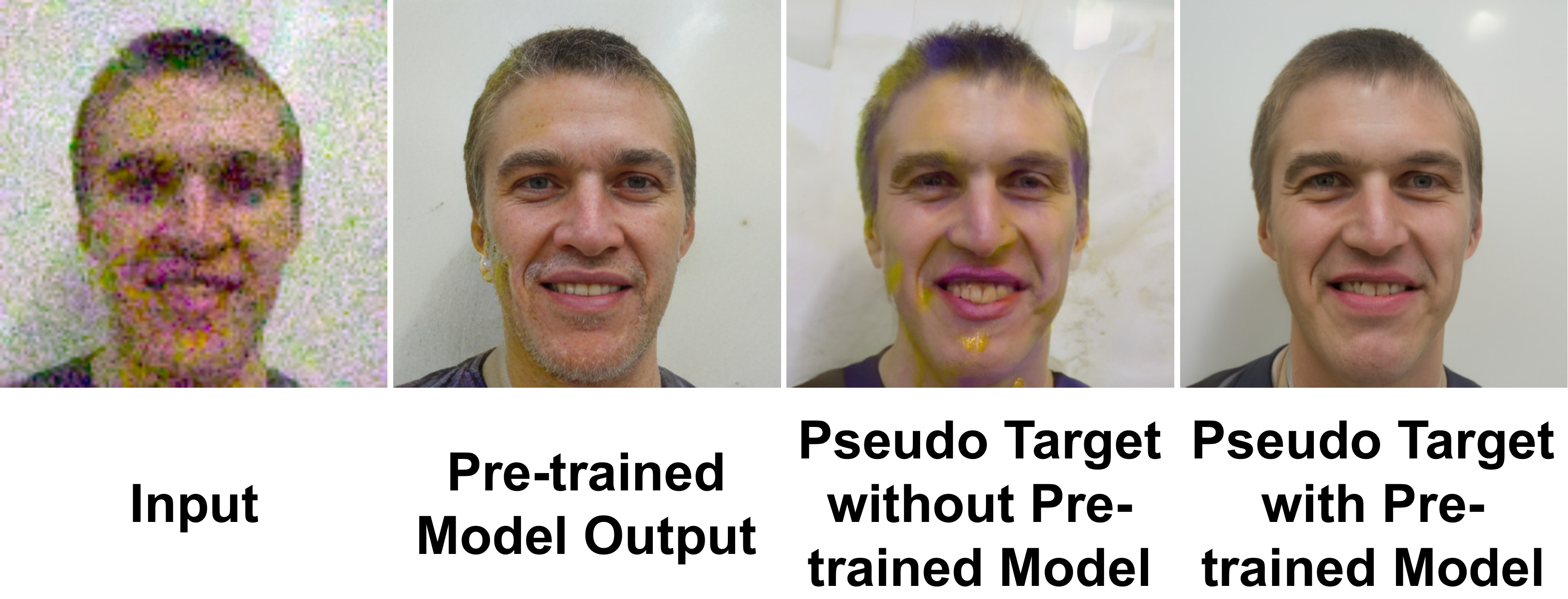}
    \caption{\textbf{Pseudo target quality without pre-trained restoration model.} We show the pseudo targets with and without running a pre-trained CodeFormer before applying target generation.}
    \label{fig:without_restoration_model}
\end{figure}

\begin{figure*}[ht]
\begin{center}
\includegraphics[width=1.0\linewidth]{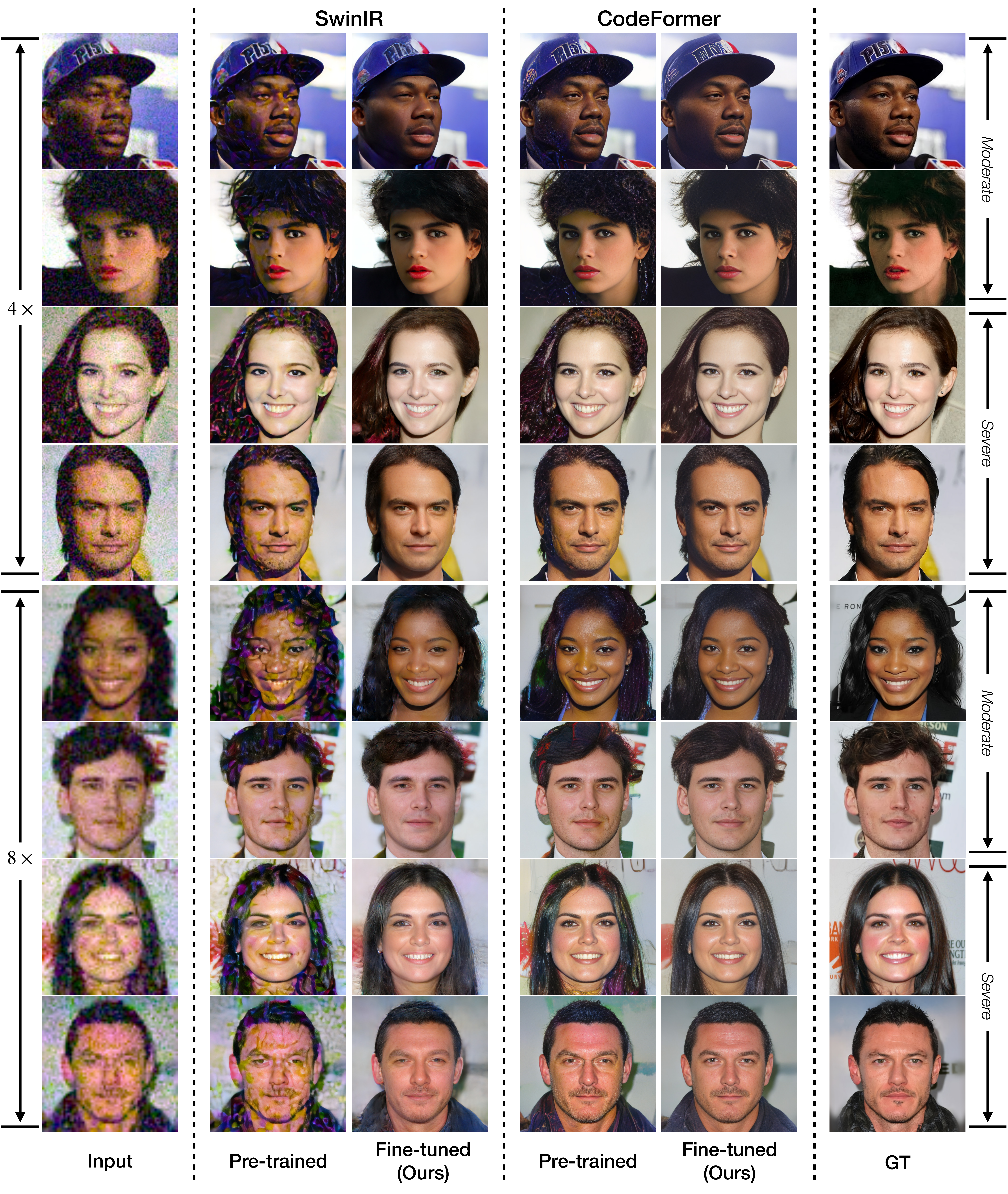}
\end{center}
\vspace{-10pt}
\caption{Additional qualitative comparison between pre-trained and fine-tuned models at different degradation levels (\textbf{zoom in for details}).}
\label{fig:finetuning_appendix}
\vspace{-1.0em}
\end{figure*}

As shown in the qualitative results in the main paper and in~\cref{fig:finetuning_appendix} in this supplementary material, a pre-trained model effectively removes some artifacts from the inputs, while preserving good fidelity of the image content. Better initial images for pseudo target generation leads to better pseudo target quality, as shown in~\cref{fig:without_restoration_model}. The pre-trained CodeFormer is able to produce a cleaner image for the pseudo target generation process, despite that its output still contains artifacts. These artifacts are cleaned up by our pseudo target generation process. Using the better images also benefit the low-frequency constrained denoising process as we will have a more accurate and less noisy constraint applied in target generation. We also provide quantitative results for this behaviour in~\cref{tab:with_vs_without_codeformer_targets}, where the pseudo targets generated from a pre-trained CodeFormer's outputs are consistently better than the ones generated directly from degraded inputs, and ultimately make the fine-tuned CodeFormer models better (see~\cref{tab:with_vs_without_codeformer_fine-tuning}) at all noise levels and downsampling levels.

\subsection{Pseudo Targets Fidelity - Quality Trade-off}

\begin{figure}[t!]
    \centering
    \includegraphics[width=\linewidth, trim={0cm 0.1cm 0cm 0.2cm},clip]{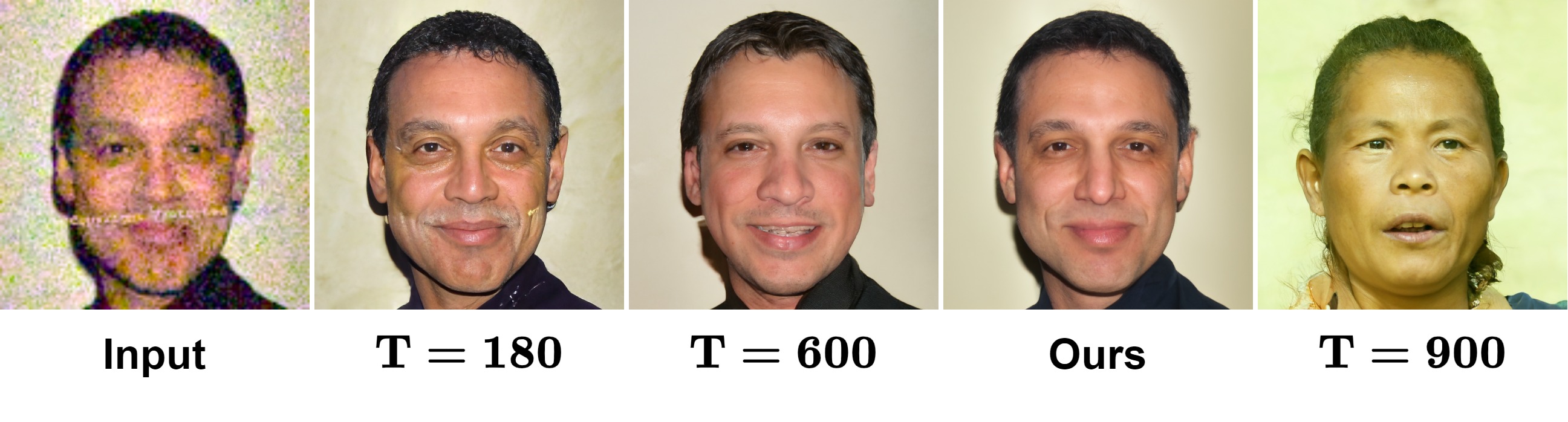}
    \caption{\textbf{Pseudo target quality vs. fidelity.} We show the pseudo targets generated with unconditional denosing up to timestep of 900 and with our optimal timestep setup (combination of low frequency constrained denoising and unconditional denoising). When the timestep is large enough, the generated target becomes a completely different face.}
    \label{fig:pseudo_target_quality_vs_fidelity}
\end{figure}

\begin{table}[t!]
  \centering
  \resizebox{\columnwidth}{!}{
      \begin{tabular}{l|c c c c c c}
        \toprule
Setup &   PSNR$\uparrow$ & SSIM$\uparrow$ & LPIPS$\downarrow$ & FID-GT$\downarrow$ & FID-FFHQ$\downarrow$\\
    \midrule
$T=180$ & 22.83 & 0.5981 & 0.4498 & 47.41 & 71.17 \\
$T=600$ & 20.11 & 0.5746 & 0.4857 & 41.94 & 59.43 \\
$T=900$ & 12.96 & 0.4066 & 0.7017 & 66.98 & 49.49 \\
    \midrule
Ours & 23.15 & 0.6507 & 0.4364 & 37.42 & 67.40 \\
        \bottomrule
      \end{tabular}
    }
  \caption{\textbf{Pseudo targets fidelity vs. quality.} We compare the  pseudo targets in different target generation setups on 4$\times$ downsampling data at \textit{severe} noise level. The first three rows correspond to unconditional denoising with different starting timesteps, and the last row corresponds to our target generation setup in this work. Note that specifically in this table, FID-GT refers to the FID score against the statistics of the ground-truth image set (measures a combination of fidelity and quality, while FID-FFHQ refers to the FID score against the FFHQ dataset statistics (measures overall face image quality and not fidelity). As described in the Metrics section of the main paper, we report FID-GT as the FID score for evaluations on synthetic datasets in all the other tables of this work, which aligns better with the targets of interest when available.}
  \label{tab:fidelity_vs_quality_targets}
\end{table}

Our pseudo targets may not be perfectly aligned with inputs in terms of fidelity, because the unconditional diffusion model is trained for generation rather than for restoration. The low-frequency constraint during the diffusion denoising process does not ensure exact matching in high-frequency details. Such process improves the image quality at the expense of fidelity. The property of an unconditional diffusion model is that, starting the denoising process at higher timesteps (more Gaussian noise injected into the image) will apply more distortion to the content of the denoised image, which leads to worse fidelity but higher overall image quality (higher FID-FFHQ score in~\cref{tab:fidelity_vs_quality_targets}). This means that one can generate pseudo targets with great image quality while completely losing the identity information of the face to be restored, as shown in~\cref{fig:pseudo_target_quality_vs_fidelity}. Starting at timestep of 900 for a 1,000 steps diffusion model provides pseudo target of another person. Such case will be catestropic for the later fine-tuning because the networks are optimized with image-level losses ($L1$ and LPIPS losses), which has strict requirements in terms of fidelity between target and input. As shown in~\cref{tab:fidelity_vs_quality_targets_finetuning}, fine-tuning with inconsistent targets ($T=900$) lead to very poor results. Our setup with constrained denoising process finds a good balance between fidelity and quality for the pseudo targets, which is an essential part of its success.

\subsection{Improved DifFace and DiffBIR}

DifFace~\cite{yue2022difface} and DiffBIR~\cite{lin2023diffbir} are two diffusion-based approaches that apply a pre-processing restoration model first to remove some artifacts on the images before passing images to the diffusion models. Both methods use the same pre-trained restoration model, a SwinIR~\cite{liang2021swinir} that is pre-trained on the FFHQ dataset~\cite{ffhq} with MSE loss only. The motivation behind this choice of loss is that this model would output blurry faces, but with good fidelity. The diffusion models are then responsible for generating high-frequency details. However, these two methods would fail if this pre-trained SwinIR fails on inputs with out-of-distribution degradation. In~\cref{tab:improved_difface_diffbir}, we show the improvements of DifFace and DiffBIR after using our fine-tuning pipeline. Note that this fine-tuned SwinIR is based on the pre-trained SwinIR with MSE loss only, not from the pre-trained SwinIR we used for other experiments in this work. The results show that with our fine-tuning approach, one can improve methods that rely on a restoration model for pre-processing.


\begin{table}[t!]
  \centering
  \resizebox{\columnwidth}{!}{
      \begin{tabular}{l|c c c c c}
        \toprule
Pseudo Target Setup &   PSNR$\uparrow$ & SSIM$\uparrow$ & LPIPS$\downarrow$ & FID$\downarrow$\\
    \midrule
$T=180$ & 22.81 & 0.5849 & 0.4294 & 46.00 \\
$T=600$ & 21.45 & 0.5487 & 0.4581 & 48.72 \\
$T=900$ & 16.90 & 0.3591 & 0.5673 & 104.54 \\
    \midrule
Ours & 22.85 & 0.6036 & 0.4258 & 41.21 \\
        \bottomrule
      \end{tabular}
    }
  \caption{\textbf{Effects of pseudo targets fidelity vs. quality in fine-tuning.} We compare the results of fine-tuning using pseudo targets in different target generation setups on 4$\times$ downsampling data at \textit{severe} noise level. The first three rows correspond to targets with unconditional denoising with different starting timesteps, and the last row corresponds to our targets generation setup in this work.}
  \label{tab:fidelity_vs_quality_targets_finetuning}
\end{table}

\section{Limitations}
\label{sec:limitations}

Our work relies on a robust pre-trained diffusion model, and the diffusion model must be pre-trained to generate images of the same category as the low-quality inputs. This similarity requirement is only in terms of the image categories (e.g. faces, animals, natural images), while the types of degradations in the low-quality inputs are independent from the pre-trained diffusion model. As part of the inputs to our pipeline, a set of unpaired low-quality images is needed for the purpose of fine-tuning the pre-trained model. As shown in~\cref{tab:num_imgs_swinir}, despite being able to improve the pre-trained model's performance, having a small set of images with only $\sim$20 images can potentially lead to over-fitting for certain models~\cite{liang2021swinir}, which deteriorates restoration quality. In addition, one may need to manually modify the downsampling factor $N$ for low pass filter based on the quality of the pre-trained model's restoration (more severe artifacts require larger $N$) although we've shown that using timesteps of $K=600$ and $L=360$ are robust to various quality of restoration. If the pre-trained restoration model fails completely, our pseudo target generation will not output feasible target either as it depends on the restoration model's output. We suggest the users to directly feed input image to our pseudo target generation pipeline in this extreme case.

\begin{table}[t!]
  \centering
  \resizebox{\columnwidth}{!}{
      \begin{tabular}{c|c|c c c c}
        \toprule
    Model & With which SwinIR? & PSNR$\uparrow$ & SSIM$\uparrow$ & LPIPS$\downarrow$ & FID$\downarrow$ \\
    \midrule

    \multirow{2}{*}{DifFace~\cite{yue2022difface}} & Pre-trained & 21.21 & 0.5645 & 0.5504 & 84.99 \\
     & Fine-tuned & \textbf{23.11} & \textbf{0.6210} & \textbf{0.4438} & \textbf{60.38} \\
    \midrule
    \midrule
    \multirow{2}{*}{DiffBIR~\cite{lin2023diffbir}} & Pre-trained & 21.59 & 0.5371 & 0.6172 & 100.81 \\
     & Fine-tuned & \textbf{22.89} & \textbf{0.5769} & \textbf{0.4750} & \textbf{55.78}\\
        \bottomrule
      \end{tabular}}
  \vspace{5pt}
  \caption{\textbf{Improved DifFace~\cite{yue2022difface} and DiffBIR~\cite{lin2023diffbir} with fine-tuned SwinIR.} We show the improvements gained for DifFace and DiffBIR by replacing their pre-trained SwinIR with the fine-tuned counterpart on our synthetic datasets.}
  \label{tab:improved_difface_diffbir}
\end{table}

\section{Potential Negative Impact}
\label{sec:negative_effects}

In this work, we use the FFHQ dataset~\cite{ffhq} to pre-train and use a subset of the CelebA-HQ~\cite{celebahq} to fine-tune our models. The two datasets are publicly available, where FFHQ consists of 70,000 high-quality face images crawled from Flickr and CelebA-HQ contains high-resolution version of the face images of different celebrities from CelebA dataset~\cite{celeba}. Our trained models will inherit the existing bias in these face datasets, particularly the imbalance in the distribution of gender, ethnicity, and age~\cite{huber2024bias}. This could potentially limit the model's performance on certain underrepresented groups of population. It also leads to perpetuation and amplification of societal biases within the current datasets. A large-scale face image dataset that is more balanced and diverse is needed for future research. In our Wider-Test-200 dataset, we make the effort to improve the diversity of the faces in our test sets by including testing images from different gender, ethnicity, and age groups. In addition, the misuse of our pipeline will pose ethical issues on potential personal privacy breach and illegal manipulation of face images.


\begin{table*}[ht]
  \centering
  \addtolength{\tabcolsep}{1pt}
      \begin{tabular}{l|c|c c c c|c c c c}
        \toprule
          & Noise & \multicolumn{4}{c|}{4$\times$ Downsampling} & \multicolumn{4}{c}{8$\times$ Downsampling} \\
          & Level & PSNR$\uparrow$ & SSIM$\uparrow$ & LPIPS$\downarrow$ & FID$\downarrow$ & PSNR$\uparrow$ & SSIM$\uparrow$ & LPIPS$\downarrow$ & FID$\downarrow$ \\
        \midrule
 Pre-trained        &     moderate       & 21.28 & 0.5744 & 0.5446 &  74.12 & 21.28 & 0.5744 & 0.5446 & 99.44\\ 
 Pseudo targets        &     moderate       & 22.89 & 0.6317 & 0.4444 &  \textbf{41.13} & 21.23 & 0.5819 & 0.5168 & 77.89 \\ 
 Fine-tuned             &     moderate    & \textbf{24.75} & \textbf{0.6676} &  \textbf{0.3853} & 41.42   &  \textbf{23.28} & \textbf{0.6206} & \textbf{0.4348} & \textbf{68.25} \\ 
 \midrule
 \midrule
 Pre-trained        &     severe       & 20.92 & 0.5444 & 0.5842 &  120.44 & 19.00 & 0.4813 & 0.6435 & 152.90 \\ 
 Pseudo targets        &     severe       & 21.24 & 0.5871 & 0.4950 &  \textbf{47.47} & 19.53 & 0.5300 & 0.5822 & \textbf{86.01} \\ 
 Fine-tuned              &     severe    & \textbf{23.41} & \textbf{0.6284} & \textbf{0.4156} & 49.46  &  \textbf{21.91} & \textbf{0.5720} & \textbf{0.4821} & 115.97\\ 
         \bottomrule
      \end{tabular}
  \vspace{5pt}
  \caption{\textbf{SwinIR's improvements after fine-tuning.} We show the effectiveness of our approach on a pre-trained SwinIR on all four of our synthetic data setups.}
  \label{tab:swinir_appendix}
\end{table*}

\begin{table*}[ht]
  \centering
  \addtolength{\tabcolsep}{1.5pt}
      \begin{tabular}{l|c c c c|c c c c}
        \toprule
         & \multicolumn{4}{c|}{4$\times$ Downsampling} & \multicolumn{4}{c}{8$\times$ Downsampling} \\
          & PSNR$\uparrow$ & SSIM$\uparrow$ & LPIPS$\downarrow$ & FID$\downarrow$ & PSNR$\uparrow$ & SSIM$\uparrow$ & LPIPS$\downarrow$ & FID$\downarrow$ \\
        \midrule
DifFace~\cite{yue2022difface}                   & 23.32 & 0.6399 & 0.4467 & 43.48 & 21.05 & 0.5649 & 0.5564 & 87.91  \\
 DiffBIR~\cite{lin2023diffbir}                 & \textbf{23.95} & 0.6185 & 0.5217 & 64.33 & 21.30  & 0.5356 & 0.6134 & 92.02  \\
PG-Diff~\cite{yang2023pgdiff}                  & 23.41 & \textbf{0.6515} & 0.4235 & 41.30  & 20.68 & 0.5799 & 0.4992 & 87.53  \\
 DR2~\cite{wang2023dr2}                    & 22.01 & 0.6168 & 0.4400   & 55.78 & 21.35 & 0.5962 & 0.4548 & 51.58  \\
 Our pseudo targets & 23.08 & 0.6407 & \textbf{0.4241} & \textbf{36.45} & \textbf{22.17} & \textbf{0.6185} & \textbf{0.4493} & \textbf{42.00}     \\
\midrule
\midrule
 GFPGAN~\cite{gfpgan}                    & 24.12 & \textbf{0.6454} & 0.4385 & 45.76 & 21.90  & 0.5629 & 0.5013 & 67.94  \\
 VQFR~\cite{vqfr}                      & 21.78 & 0.5372 & 0.4711 & 83.84 & 20.22 & 0.4919 & 0.5184 & 104.12 \\
 CodeFormer~\cite{codeformer}    & \textbf{24.31} & 0.6335 & \textbf{0.4007} & \textbf{40.66} & 22.19 & 0.5716 & 0.4420  & 51.91  \\
 CodeFormer + Ours       & 23.20  & 0.6138 & 0.4117 & 41.74 & \textbf{22.28} & \textbf{0.5848} & \textbf{0.4290}  & \textbf{41.72}  \\
         \bottomrule
      \end{tabular}
  \vspace{5pt}
  \caption{\textbf{CodeFormer results on data with \textbf{moderate} noise level.} We compare our pseudo targets and fine-tuned results with pre-trained CodeFormer and other baselines. Top rows: diffusion-dependent models at test time. Bottom rows: diffusion-free models at test time.}
  \label{tab:moderate_codeformer_appendix}
\end{table*}

\begin{table*}[ht]
  \centering
  \addtolength{\tabcolsep}{1.5pt}
      \begin{tabular}{l|c c c c|c c c c}
        \toprule
         & \multicolumn{4}{c|}{4$\times$ Downsampling} & \multicolumn{4}{c}{8$\times$ Downsampling} \\
          & PSNR$\uparrow$ & SSIM$\uparrow$ & LPIPS$\downarrow$ & FID$\downarrow$ & PSNR$\uparrow$ & SSIM$\uparrow$ & LPIPS$\downarrow$ & FID$\downarrow$ \\
        \midrule
DifFace~\cite{yue2022difface}                   & 21.46 & 0.5731 & 0.5396 & 75.78  & 18.99 & 0.4800   & 0.6589 & 132.81 \\
 DiffBIR~\cite{lin2023diffbir}                 & 21.83 & 0.5380  & 0.6262 & 86.05  & 19.26 & 0.4562 & 0.7076 & 160.85 \\
PG-Diff~\cite{yang2023pgdiff}                  & 21.82 & 0.6012 & 0.4923 & 64.82  & 18.31 & 0.4879 & 0.5976 & 152.10  \\
 DR2~\cite{wang2023dr2}                    & 20.13 & 0.5631 & 0.4818 & 53.16  & 19.32 & 0.5353 & 0.5082 & 58.30   \\
 Our pseudo targets & \textbf{23.15} & \textbf{0.6507} & \textbf{0.4364} & \textbf{37.42}  & \textbf{21.14} & \textbf{0.5902} & \textbf{0.4918} & \textbf{50.59}  \\
\midrule
\midrule
 GFPGAN~\cite{gfpgan}                    & 22.89 & 0.5999 & 0.4807 & 56.07  & 20.27 & 0.4937 & 0.5582 & 97.86  \\
 VQFR~\cite{vqfr}                      & 20.16 & 0.4709 & 0.5318 & 114.00    & 18.54 & 0.4161 & 0.5950  & 142.11 \\
 CodeFormer~\cite{codeformer}    & \textbf{22.90}  & 0.5810  & 0.4420  & 53.02  & 20.60  & 0.5108 & 0.4938 & 72.16  \\
 CodeFormer + Ours & 22.85 & \textbf{0.6036} & \textbf{0.4258} & \textbf{41.21} & \textbf{21.38} & \textbf{0.5514} & \textbf{0.4589} & \textbf{46.66}  \\
         \bottomrule
      \end{tabular}
  \caption{\textbf{CodeFormer results on data with \textbf{severe} noise level.} We compare our pseudo targets and fine-tuned results with pre-trained CodeFormer and other baselines. Top rows: diffusion-dependent models at test time. Bottom rows: diffusion-free models at test time.}
  \label{tab:severe_codeformer_appendix}
  \vspace{1em}
\end{table*}

\begin{table*}[ht]
  \centering
  \addtolength{\tabcolsep}{1pt}
      \begin{tabular}{c|c c c c|c c c c}
        \toprule
          Noise & \multicolumn{4}{c|}{4$\times$ Downsampling} & \multicolumn{4}{c}{8$\times$ Downsampling} \\
          Level & PSNR$\uparrow$ & SSIM$\uparrow$ & LPIPS$\downarrow$ & FID$\downarrow$ & PSNR$\uparrow$ & SSIM$\uparrow$ & LPIPS$\downarrow$ & FID$\downarrow$ \\ \midrule
moderate & 26.79 & 0.7201 & 0.3302 & 32.69 & 24.65 & 0.6618 & 0.3722 & 37.52 \\
severe & 25.78 & 0.6894 & 0.3522 & 34.99 & 23.75 & 0.6447 & 0.3908 & 40.16 \\
         \bottomrule
      \end{tabular}
  \caption{\textbf{SwinIR results using GT images as pseudo targets.} We show the results of fine-tuning a pre-trained SwinIR using GT clean images on all four of our synthetic data setups (supervised fine-tuning).}
  \label{tab:swinir_gt_targets_appendix}
\end{table*}

\begin{table*}[ht]
  \centering
  \addtolength{\tabcolsep}{1pt}
      \begin{tabular}{c|c c c c|c c c c}
        \toprule
  Noise & \multicolumn{4}{c|}{4$\times$ Downsampling} & \multicolumn{4}{c}{8$\times$ Downsampling} \\
  Level & PSNR$\uparrow$ & SSIM$\uparrow$ & LPIPS$\downarrow$ & FID$\downarrow$ & PSNR$\uparrow$ & SSIM$\uparrow$ & LPIPS$\downarrow$ & FID$\downarrow$ \\ \midrule
moderate & 24.53 & 0.6565 & 0.3598 & 36.57 & 23.38 & 0.6169 & 0.3881 & 39.06 \\
severe & 24.02 & 0.6369 & 0.3755 & 37.00 & 22.73 & 0.5978 & 0.4063 & 40.98 \\
         \bottomrule
      \end{tabular}
  \caption{\textbf{CodeFormer results using GT images as pseudo targets.} We show the results of fine-tuning a pre-trained CodeFormer using GT clean images on all four of our synthetic data setups (supervised fine-tuning).}
  \label{tab:codeformer_gt_targets_appendix}
\end{table*}

\begin{table*}[ht]
  \centering
  \addtolength{\tabcolsep}{3.5pt}
      \begin{tabular}{c |c c c | c c c c }
        \toprule
 \multicolumn{1}{c|}{\begin{tabular}[c]{@{}c@{}}Low Freq. \\ Constraint\end{tabular}} & \multicolumn{1}{c}{\begin{tabular}[c]{@{}c@{}}Low Pass \\ Filter's $N$\end{tabular}} & \multicolumn{1}{c}{\begin{tabular}[c]{@{}c@{}} Timestep \\ $K$\end{tabular}} & \multicolumn{1}{c|}{\begin{tabular}[c]{@{}c@{}} Timestep \\ $L$\end{tabular}} & PSNR$\uparrow$ & SSIM$\uparrow$ & LPIPS$\downarrow$ & FID$\downarrow$ \\
\midrule
\cmark & 8  & 600 & 360 & \textcolor{blue}{23.47} & \textcolor{red}{0.6564} & \textcolor{red}{0.4381} & 43.01 \\
\cmark & \textbf{16} & \textbf{600} & \textbf{360} & 22.89 & 0.6317 & \textcolor{blue}{0.4444} & \textcolor{blue}{41.13} \\
\cmark & 32 & 600 & 360 & 22.05 & 0.6127 & 0.4504 & \textcolor{red}{40.92} \\
\cmark & 16 & 600 & 0   & 22.73 & 0.6114 & 0.4991 & 58.10 \\
\cmark & 16 & 600 & 180 & 22.85 & 0.6204 & 0.4786 & 55.03 \\
\cmark & 16 & 900 & 360 & 22.59 & 0.6277 & 0.4512 & 43.20 \\
\xmark & -  & 600 & -   & 20.53 & 0.5874 & 0.4717 & 43.51 \\
\xmark & -  & 360 & -   & 23.04 & 0.6319 & 0.4568 & 51.84 \\
\xmark & -  & 180 & -   & \textcolor{red}{23.52} & \textcolor{blue}{0.6446} & 0.4872 & 71.18 \\
        \bottomrule
      \end{tabular}
  \vspace{5pt}
  \caption{\textbf{Quantitative ablation study on the low pass downsampling factor and timestep choices for SwinIR pseudo targets.} We compare the results of the SwinIR pseudo targets with different timesteps choices ($K$ and $L$) and low pass filter downsampling parameters ($N$) on 4$\times$ downsampling data at  \textit{moderate} noise level. \textcolor{red}{Red} and \textcolor{blue}{blue} indicate the best and the second best results. \textbf{Bold} indicates our selection.}
  \label{tab:swinir_targets_ablation}
\end{table*}
\begin{table*}[ht]
  \centering
  \addtolength{\tabcolsep}{3.5pt}
      \begin{tabular}{c |c c c | c c c c }
        \toprule
 \multicolumn{1}{c|}{\begin{tabular}[c]{@{}c@{}}Low Freq. \\ Constraint\end{tabular}} & \multicolumn{1}{c}{\begin{tabular}[c]{@{}c@{}}Low Pass \\ Filter's $N$\end{tabular}} & \multicolumn{1}{c}{\begin{tabular}[c]{@{}c@{}} Timestep \\ $K$\end{tabular}} & \multicolumn{1}{c|}{\begin{tabular}[c]{@{}c@{}} Timestep \\ $L$\end{tabular}} & PSNR$\uparrow$ & SSIM$\uparrow$ & LPIPS$\downarrow$ & FID$\downarrow$ \\
\midrule
\cmark & 4  & 600 & 360 & 23.37 & \textcolor{red}{0.6694} & 0.4346 & 40.85 \\
\cmark & \textbf{8}  & \textbf{600} & \textbf{360} & 23.15 & 0.6507 & \textcolor{red}{0.4364} & \textcolor{red}{37.42} \\
\cmark & 16 & 600 & 360 & 22.59 & 0.6285 & 0.4458 & 37.81 \\
\cmark & 8  & 600 & 0   & \textcolor{blue}{23.54} & 0.6409 & 0.4506 & 54.58 \\
\cmark & 8  & 600 & 180 & \textcolor{red}{23.59} & \textcolor{blue}{0.6543} & \textcolor{blue}{0.4368} & 41.28 \\
\cmark & 8  & 900 & 360 & 22.40 & 0.6266 & 0.4511 & 38.99 \\
\xmark & -  & 600 & -   & 20.11 & 0.5746 & 0.4857 & 41.94 \\
\xmark & -  & 360 & -   & 22.65 & 0.6170 & 0.4440 & \textcolor{blue}{37.45} \\
\xmark & -  & 180 & -   & 22.83 & 0.5981 & 0.4498 & 47.41 \\
        \bottomrule
      \end{tabular}
  \vspace{5pt}
  \caption{\textbf{Quantitative ablation study on low pass downsampling factor and timestep choices for CodeFormer pseudo targets.} We compare the results of CodeFormer pseudo targets with different timesteps choices ($K$ and $L$) and low pass filter downsampling parameters ($N$) on 4$\times$ downsampling data at  \textit{severe} noise level. \textcolor{red}{Red} and \textcolor{blue}{blue} indicate the best and the second best results. \textbf{Bold} indicates our selection.}
  \label{tab:codeformer_targets_ablation}
\end{table*}
\begin{table*}[ht]
  \centering
  \addtolength{\tabcolsep}{3.5pt}
      \begin{tabular}{c |c c c | c c c c }
        \toprule
 \multicolumn{1}{c|}{\begin{tabular}[c]{@{}c@{}}Low Freq. \\ Constraint\end{tabular}} & \multicolumn{1}{c}{\begin{tabular}[c]{@{}c@{}}Low Pass \\ Filter's $N$\end{tabular}} & \multicolumn{1}{c}{\begin{tabular}[c]{@{}c@{}} Timestep \\ $K$\end{tabular}} & \multicolumn{1}{c|}{\begin{tabular}[c]{@{}c@{}} Timestep \\ $L$\end{tabular}} & PSNR$\uparrow$ & SSIM$\uparrow$ & LPIPS$\downarrow$ & FID$\downarrow$ \\
\midrule
\cmark & 8  & 600 & 360 & 25.06 & \textcolor{blue}{0.6782} & \textcolor{red}{0.3803} & \textcolor{red}{40.71} \\
\cmark & \textbf{16} & \textbf{600} & \textbf{360} & 24.75 & 0.6676 & \textcolor{blue}{0.3853} & \textcolor{blue}{41.42} \\
\cmark & 32 & 600 & 360 & 24.33 & 0.6580 & 0.3943 & 42.56 \\
\cmark & 16 & 600 & 0   & 24.68 & 0.6657 & 0.4183 & 57.17 \\
\cmark & 16 & 600 & 180 & 24.70 & 0.6668 & 0.4080 & 54.01 \\
\cmark & 16 & 900 & 360 & 24.26 & 0.6567 & 0.3919 & 42.54 \\
\xmark & -  & 600 & -   & 23.72 & 0.6422 & 0.4047 & 45.40 \\
\xmark & -  & 360 & -   & \textcolor{blue}{25.09} & 0.6763 & 0.3896 & 43.72 \\
\xmark & -  & 180 & -   & \textcolor{red}{25.21} & \textcolor{red}{0.6832} & 0.4146 & 54.63 \\
        \bottomrule
      \end{tabular}
  \vspace{5pt}
  \caption{\textbf{Quantitative ablation study on the low pass downsampling factor and timestep choices for SwinIR finetuning.} We compare the results of the SwinIR pseudo targets with different timesteps choices ($K$ and $L$) and low pass filter downsampling parameters ($N$) on 4$\times$ downsampling data at  \textit{moderate} noise level. \textcolor{red}{Red} and \textcolor{blue}{blue} indicate the best and the second best results. \textbf{Bold} indicates our selection.}
  \label{tab:swinir_targets_ablation_finetuning}
\end{table*}
\begin{table*}[ht]
  \centering
  \addtolength{\tabcolsep}{3.5pt}
      \begin{tabular}{c |c c c | c c c c }
        \toprule
 \multicolumn{1}{c|}{\begin{tabular}[c]{@{}c@{}}Low Freq. \\ Constraint\end{tabular}} & \multicolumn{1}{c}{\begin{tabular}[c]{@{}c@{}}Low Pass \\ Filter's $N$\end{tabular}} & \multicolumn{1}{c}{\begin{tabular}[c]{@{}c@{}} Timestep \\ $K$\end{tabular}} & \multicolumn{1}{c|}{\begin{tabular}[c]{@{}c@{}} Timestep \\ $L$\end{tabular}} & PSNR$\uparrow$ & SSIM$\uparrow$ & LPIPS$\downarrow$ & FID$\downarrow$ \\
\midrule
\cmark & 4  & 600 & 360 & 22.87 & \textcolor{red}{0.6055} & \textcolor{red}{0.4249} & \textcolor{red}{40.71} \\
\cmark & \textbf{8}  & \textbf{600} & \textbf{360} & 22.85 & 0.6036 & 0.4258 & \textcolor{blue}{41.21} \\
\cmark & 16 & 600 & 360 & 22.66 & 0.5941 & 0.4312 & 42.30 \\
\cmark & 8  & 600 & 0   & \textcolor{red}{22.96} & 0.5927 & 
\textcolor{blue}{0.4296} & 46.73 \\
\cmark & 8  & 600 & 180 & \textcolor{blue}{22.94} & 0.6017 & 0.4256 & 42.27 \\
\cmark & 8  & 900 & 360 & 22.88 & \textcolor{blue}{0.6049} & 0.4260 & 41.61 \\
\xmark & -  & 600 & -   & 21.45 & 0.5487 & 0.4581 & 48.72 \\
\xmark & -  & 360 & -   & 22.64 & 0.5869 & 0.4291 & 42.40 \\
\xmark & -  & 180 & -   & 22.81 & 0.5849 & 0.4294 & 46.00 \\
        \bottomrule
      \end{tabular}
  \vspace{5pt}
  \caption{\textbf{Quantitative ablation study on low pass downsampling factor and timestep choices for CodeFormer finetuning.} We compare the results of CodeFormer pseudo targets with different timesteps choices ($K$ and $L$) and low pass filter downsampling parameters ($N$) on 4$\times$ downsampling data at  \textit{severe} noise level. \textcolor{red}{Red} and \textcolor{blue}{blue} indicate the best and the second best results. \textbf{Bold} indicates our selection.}
  \label{tab:codeformer_targets_ablation_finetuning}
\end{table*}

\begin{table*}[ht]
  \centering
  \addtolength{\tabcolsep}{1pt}
      \begin{tabular}{l|c|c c c c|c c c c}
        \toprule
          & Noise & \multicolumn{4}{c|}{4$\times$ Downsampling} & \multicolumn{4}{c}{8$\times$ Downsampling} \\
          & Level & PSNR$\uparrow$ & SSIM$\uparrow$ & LPIPS$\downarrow$ & FID$\downarrow$ & PSNR$\uparrow$ & SSIM$\uparrow$ & LPIPS$\downarrow$ & FID$\downarrow$ \\
        \midrule
without & moderate & 22.96 & 0.6316 & 0.4499 & 42.44 & 21.39 & 0.5808 & 0.5406 & 70.88 \\
with & moderate & \textbf{23.08} & \textbf{0.6407} & \textbf{0.4241} & \textbf{36.45} & \textbf{22.17} & \textbf{0.6185} & \textbf{0.4493} & \textbf{42.00}     \\
        \midrule
without & severe & 21.75 & 0.5941 & 0.5228 & 59.71 & 19.90 & 0.5323 & 0.6188 & 98.49 \\
with & severe & \textbf{23.15} & \textbf{0.6507} & \textbf{0.4364} & \textbf{37.42}  & \textbf{21.14} & \textbf{0.5902} & \textbf{0.4918} & \textbf{50.59}  \\
         \bottomrule
      \end{tabular}
  \vspace{5pt}
  \caption{\textbf{Pseudo targets with vs. without running a pre-trained CodeFormer.} We compare the pseudo targets generated with and without running a pre-trained CodeFormer before the pseudo target generation process.}
  \label{tab:with_vs_without_codeformer_targets}
  \vspace{-1 em}
\end{table*}

\begin{table*}[ht]
  \centering
  \addtolength{\tabcolsep}{1pt}
      \begin{tabular}{l|c|c c c c|c c c c}
        \toprule
          & Noise & \multicolumn{4}{c|}{4$\times$ Downsampling} & \multicolumn{4}{c}{8$\times$ Downsampling} \\
          & Level & PSNR$\uparrow$ & SSIM$\uparrow$ & LPIPS$\downarrow$ & FID$\downarrow$ & PSNR$\uparrow$ & SSIM$\uparrow$ & LPIPS$\downarrow$ & FID$\downarrow$ \\
        \midrule
without & moderate & 23.10 & 0.6033 & 0.4170 & 41.23 & 21.87 & 0.5442 & 0.4556 & 52.62 \\
with & moderate & \textbf{23.20}  & \textbf{0.6138} & \textbf{0.4117} & \textbf{41.74} & \textbf{22.28} & \textbf{0.5848} & \textbf{0.4290}  & \textbf{41.72} \\
        \midrule
without & severe & 22.27 & 0.5629 & 0.4505 & 46.32 & 20.63 & 0.4749 & 0.5041 & 72.73 \\
with & severe & \textbf{22.85} & \textbf{0.6036} & \textbf{0.4258} & \textbf{41.21} & \textbf{21.38} & \textbf{0.5514} & \textbf{0.4589} & \textbf{46.66}  \\
         \bottomrule
      \end{tabular}
  \vspace{5pt}
  \caption{\textbf{CodeFormer fine-tuning results using pseudo targets with vs. without running a pre-trained CodeFormer.} We compare the results of fine-tuning using pseudo targets generated with and without running a pre-trained CodeFormer before the pseudo target generation process.}
  \label{tab:with_vs_without_codeformer_fine-tuning}
\end{table*}

\begin{table*}[ht]
  \centering
  \addtolength{\tabcolsep}{1.5pt}
      \begin{tabular}{l|c c c c|c c c c}
        \toprule
          & PSNR$\uparrow$ & SSIM$\uparrow$ & LPIPS$\downarrow$ & FID$\downarrow$ & Deg.$\downarrow$ & LMD$\downarrow$ \\
        \midrule
DifFace~\cite{yue2022difface}          &     20.23 & 0.5266 & 0.5993 & 104.30 & 60.30 & 4.88 \\
 DiffBIR~\cite{lin2023diffbir}        &     20.55 & 0.4971 & 0.6669 & 123.45 & \textbf{52.15} & 4.80 \\
PG-Diff~\cite{yang2023pgdiff}        &     20.07 & 0.5445 & 0.5450 & 108.46 & 57.95 & 4.87 \\
 DR2~\cite{wang2023dr2}        &        19.73 & 0.5492 & 0.4950  & 55.73  & 67.01 & 6.37 \\
 Our pseudo targets &  \textbf{21.87} & \textbf{0.6094} & \textbf{0.4688} & \textbf{44.20}  & 57.38 & \textbf{4.35} \\
\midrule
\midrule
 GFPGAN~\cite{gfpgan}                    &  19.35 & 0.4435 & 0.5634 & 128.06 & 52.48 & 4.37 \\
 VQFR~\cite{vqfr}                      &  21.58 & 0.5468 & 0.5195 & 76.97  & 51.28 & 4.14 \\
 CodeFormer~\cite{codeformer}    &  21.75 & 0.5459 & 0.4679 & 62.59  & \textbf{49.48} & \textbf{3.83} \\
 CodeFormer + Ours      &  \textbf{22.12} & \textbf{0.5775} & \textbf{0.4424} & \textbf{43.94}  & 53.88 & 4.11 \\ 
         \bottomrule
      \end{tabular}
  \caption{\textbf{CodeFormer results of facial recognition based metrics on data with \textbf{severe} noise level.} We compare our pseudo targets and fine-tuned results with pre-trained CodeFormer and other baselines on metrics including Deg.~\cite{gfpgan} and LMD~\cite{vqfr}. Top rows: diffusion-dependent models at test time. Bottom rows: diffusion-free models at test time.}
  \label{tab:facial_recognition_metrics}
\end{table*}

\begin{table*}[ht]
  \centering
  \addtolength{\tabcolsep}{1pt}
      \begin{tabular}{c|c c c c|c c c c}
        \toprule
           & \multicolumn{4}{c|}{4$\times$ Downsampling} & \multicolumn{4}{c}{8$\times$ Downsampling} \\
           & PSNR$\uparrow$ & SSIM$\uparrow$ & LPIPS$\downarrow$ & FID$\downarrow$ & PSNR$\uparrow$ & SSIM$\uparrow$ & LPIPS$\downarrow$ & FID$\downarrow$ \\ \midrule
Pre-trained & 25.32 & 0.6909 & 0.4173 & 46.34 & 23.12 & 0.6341 & 0.4752 & 66.43 \\
Fine-tuned & \textbf{25.37} & \textbf{0.6872} & \textbf{0.3658} & \textbf{39.03} & \textbf{23.75} & \textbf{0.6412} & \textbf{0.4063} & \textbf{48.87} \\
         \bottomrule
      \end{tabular}
\caption{\textbf{SwinIR's results on data with \textit{mild} degradations.} We show the effectiveness of our approach on a pre-trained SwinIR on inputs with mild degradations.}
  \label{tab:swinir_mild_degradation}
\end{table*}

\begin{table*}[ht]
  \centering
  \addtolength{\tabcolsep}{1.5pt}
      \begin{tabular}{l|c c}
        \toprule
 & Number of Parameters~$\downarrow$ & Inference Time~$\downarrow$ \\
        \midrule
DifFace~\cite{yue2022difface}                   & 159.59M & 4.0053s \\
 DiffBIR~\cite{lin2023diffbir}                 & 1666.75M  & 9.1807s \\
PG-Diff~\cite{yang2023pgdiff}                  & 159.59M & 15.9534s \\
 DR2~\cite{wang2023dr2}                    & 93.56M & 1.0812s   \\
\midrule
\midrule
 GFPGAN~\cite{gfpgan}                    & 76.21M & 0.0249s  \\
 VQFR~\cite{vqfr}                      & 76.56M & 0.1392s \\
 One step of diffusion model~\cite{ho2020denoising}        & 159.59M & 0.0402s \\
 SwinIR~\cite{liang2021swinir} + Ours    &  15.79M & 0.0311s \\
 CodeFormer~\cite{codeformer} + Ours   &  94.11M & 0.0274s \\
         \bottomrule
      \end{tabular}
  \vspace{5pt}
  \caption{\textbf{Memory and inference time comparison.} We compare the memory and inference time of the baseline methods with the model architectures we used in our pipeline. Note that the one step of the diffusion model refers to the time it performs one denoising step.}
  \label{tab:runtime_memory_comparison}
\end{table*}

\clearpage


\begin{figure}[t]
\begin{algorithm}[H]
\caption{Generating pseudo targets (ours)}
\begin{algorithmic}
\State \textbf{Input}: low-quality restoration output $y_0=\mathcal{R}(y)$, low-pass filter $\phi_{N}$, pre-defined timesteps $L$ and $K$ where $L < K < T$
\State \textbf{Output}: pseudo target $\bar{x}_{0}$ for low-quality input $y$
\State $\bar{x}_K \gets \text{sample from}~\mathcal{N}(y_K; \sqrt{\bar \alpha_K}y_{0}, (1-\bar \alpha_K)\boldsymbol{\mathrm{I}})$
\For{$t$ from $K$ to $1$}
    \State $\bar{x}_{t-1} \gets \text{sample from } p_\theta(\bar{x}_{t-1} | \bar{x}_t)$ \Comment{unconditional denoising}
    \If{$ t > L$}
        \State ${y}_{t-1} \gets \text{sample from}~\mathcal{N}({y}_{t-1}; \sqrt{\bar \alpha_{t-1}}y_{0},   (1 - \bar \alpha_{t-1})\boldsymbol{\mathrm{I}})$
        \State $\bar{x}_{t-1} \gets \bar{x}_{t-1} - \phi_N(\bar{x}_{t-1}) + \phi_N({y}_{t-1})$ \Comment{low frequency content constraint}
     \EndIf
    \EndFor
\State \Return $\bar{x}_{0}$
\end{algorithmic}
\label{alg:sampling_ours}
\end{algorithm}
\vspace{-1.0 em}
\end{figure}

\begin{figure}[t]
\begin{algorithm}[H]
\caption{DifFace~\cite{yue2022difface}}
\begin{algorithmic}
\State \textbf{Input}: output of a pre-trained restoration model $y_0=\mathcal{R}(y)$, a pre-defined timestep $K$ where $K<T$
\State \textbf{Output}: clean image $x_0$ for low-quality input $y$
\State $x_K \gets \text{sample from}~\mathcal{N}(y_K; \sqrt{\bar \alpha_K}y_{0}, (1-\bar \alpha_K)\boldsymbol{\mathrm{I}})$
\For{$t$ from $K$ to $1$}
    \State $x_{t-1} \gets \text{sample from } p_\theta(x_{t-1} | x_t)$ \Comment{unconditional denoising}
    \EndFor
\State \Return $x_{0}$
\end{algorithmic}
\label{alg:sampling_difface}
\end{algorithm}
\end{figure}

\begin{figure}[t]
\begin{algorithm}[H]
\caption{ILVR~\cite{ilvr}}
\begin{algorithmic}
\State \textbf{Input}: low-quality input $y$, low-pass filter $\phi_{N}$
\State \textbf{Output}: clean image $x_0$ for low-quality input $y$
\State $x_T \gets \text{sample from}~\mathcal{N}(\bm{0}; \boldsymbol{\mathrm{I}})$
\For{$t$ from $T$ to $1$}
    \State $x_{t-1} \gets \text{sample from } p_\theta(x_{t-1} | x_t)$ \Comment{unconditional denoising}
    \State ${y}_{t-1} \gets \text{sample from}~\mathcal{N}({y}_{t-1}; \sqrt{\bar \alpha_{t-1}}y,   (1 - \bar \alpha_{t-1})\boldsymbol{\mathrm{I}})$
    \State $x_{t-1} \gets x_{t-1} - \phi_N(x_{t-1}) + \phi_N({y}_{t-1})$
    \EndFor
\State \Return $x_{0}$
\end{algorithmic}
\label{alg:sampling_ilvr}
\end{algorithm}
\end{figure}

\begin{figure}[t]
\begin{algorithm}[H]
\caption{DDA~\cite{gao2023back}}
\begin{algorithmic}
\State \textbf{Input}: low-quality input $y$, low-pass filter $\phi_{N}$, a pre-defined timestep $K$ where $K<T$, diffusion model's noise prediction network $\boldsymbol{\epsilon}_{\theta}$, guidance weight $\boldsymbol{s}$
\State \textbf{Output}: clean image $x_0$ for low-quality input $y$
\State $x_K \gets \text{sample from}~\mathcal{N}(y_K; \sqrt{\bar \alpha_K}y, (1-\bar\alpha_K)\boldsymbol{\mathrm{I}})$
\For{$t$ from $K$ to $1$}
    \State $\hat{x}_{t-1} \gets \text{sample from} p_\theta(x_{t-1} | x_t)$ \Comment{unconditional denoising}
    \State $\hat{x}_{0} \gets ({x}_{t} - \sqrt{1-\bar{\alpha}_t}\boldsymbol{\epsilon}_{\theta}(x_t, t)) / \sqrt{\bar{\alpha}_t}$ \Comment{Estimate $x_0$ from $x_t$ directly}
    \State $x_{t-1} \gets \hat{x}_{t-1} - \boldsymbol{s}\nabla_{x_t}||\phi_N(y) - \phi_N(\hat{x}_0)||_{2}$ \Comment{gradient guidance}
    \EndFor
\State \Return $x_{0}$
\end{algorithmic}
\label{alg:sampling_dda}
\end{algorithm}
\end{figure}

\begin{figure}[t]
\begin{algorithm}[H]
\caption{PG-Diff~\cite{yang2023pgdiff}}
\begin{algorithmic}
\State \textbf{Input}: output of a pre-trained restoration model $y_0=\mathcal{R}(y)$, pre-defined timesteps $\tau$ and $K$ where $\tau < K < T$, unconditional denoising $p_\theta(\hat{x}_{t-1} | \hat{x}_t)=\mathcal{N}(\mu_{\theta}, \Sigma_{\theta})$, diffusion model's noise prediction network $\boldsymbol{\epsilon}_{\theta}$, guidance weight $\boldsymbol{s}$, number of gradient steps $G$
\State \textbf{Output}: clean image $x_0$ for low-quality input $y$
\State $x_T \gets \text{sample from}~\mathcal{N}(\bm{0}; \boldsymbol{\mathrm{I}})$
\For{$t$ from $T$ to $1$}
    \State $\mu, \Sigma \gets \mu_{\theta}(x_t, t), \Sigma_{\theta}(x_t, t)$
    \State $\hat{x}_{0} \gets ({x}_{t} - \sqrt{1-\bar{\alpha}_t}\boldsymbol{\epsilon}_{\theta}(x_t, t)) / \sqrt{\bar{\alpha}_t}$ \Comment{Estimate $x_0$ from $x_t$ directly}
    \If{$ \tau \leq t \leq K$} \Comment{multiple guidance steps}
        \Repeat 
        \State $x_{t} \gets \text{sample from } \mathcal{N}(\mu - \boldsymbol{s}\nabla_{\hat{x}_{0}}||y_0 - \hat{x}_0||_{2}^{2}, \Sigma)$
        \State $\hat{x}_{0} \gets ({x}_{t} - \sqrt{1-\bar{\alpha}_t}\boldsymbol{\epsilon}_{\theta}(x_t, t)) / \sqrt{\bar{\alpha}_t}$
        \Until{$G-1$ times}
    \EndIf
    \State $x_{t-1} \gets \text{sample from } \mathcal{N}(\mu - \boldsymbol{s}\nabla_{\hat{x}_{0}}||y_0 - \hat{x}_0||_{2}^{2}, \Sigma)$ \Comment{gradient guidance}
    \EndFor
\State \Return $x_{0}$
\end{algorithmic}
\label{alg:sampling_pgdiff}
\end{algorithm}
\end{figure}

\begin{figure}[t]
\begin{algorithm}[H]
\caption{DiffBIR~\cite{lin2023diffbir}}
\begin{algorithmic}
\State \textbf{Input}: output of a pre-trained restoration model $y_0=\mathcal{R}(y)$, unconditional denoising of a latent diffusion model $p_\theta(\hat{z}_{t-1} | \hat{z}_t)=\mathcal{N}(\mu_{\theta}, \Sigma_{\theta})$, a fine-tuned latent diffusion model's noise prediction network $\boldsymbol{\epsilon}_{\theta}$ with text prompt set to empty, latent diffusion model's encoder $E$ and decoder $D$, guidance weight $\boldsymbol{s}$
\State \textbf{Output}: clean image $x_0$ for low-quality input $y$
\State ${z}_{{y}_{0}} \gets E(y_0)$
\State $z_T \gets \text{sample from}~\mathcal{N}(\bm{0}; \boldsymbol{\mathrm{I}})$
\For{$t$ from $T$ to $1$}
    \State $\mu, \Sigma \gets \mu_{\theta}(x_t, t), \Sigma_{\theta}(x_t, t)$
    \State $\hat{z}_{0} \gets ({z}_{t} - \sqrt{1-\bar{\alpha}_t}\boldsymbol{\epsilon}_{\theta}(z_t, t)) / \sqrt{\bar{\alpha}_t}$ \Comment{Estimate $z_0$ from $z_t$ directly}
    \State $z_{t-1} \gets \text{sample from } \mathcal{N}(\mu - \boldsymbol{s}\nabla_{\hat{x}_{0}}||{z}_{{y}_{0}} - \hat{z}_0||_{2}^{2}, \Sigma)$ \Comment{gradient guidance}
    \EndFor
\State $x_0 \gets D(z_0)$
\State \Return $x_0$
\end{algorithmic}
\label{alg:sampling_diffbir}
\end{algorithm}
\end{figure}

\begin{figure}[t]
\begin{algorithm}[H]
\caption{DR2~\cite{wang2023dr2}}
\begin{algorithmic}
\State \textbf{Input}: low-quality input $y$, low-pass filter $\phi_{N}$, a pre-trained face restoration model $f$ for post-processing, pre-defined timesteps $\tau$ and $K$ where $\tau < K < T$, diffusion model's noise prediction network $\boldsymbol{\epsilon}_{\theta}$, downsampling factor $r=2$
\State \textbf{Output}: clean image $x_0$ for low-quality input $y$
\State $y_0 \gets (y)_{\downarrow_{r}}$ \Comment{Downsampling by a factor of $r$}
\State $\hat{x}_K \gets \text{sample from}~\mathcal{N}(y_K; \sqrt{\bar \alpha_K}y_{0}, (1-\bar \alpha_K)\boldsymbol{\mathrm{I}})$
\For{$t$ from $K$ to $(\tau+1)$}
    \State $\hat{x}_{t-1} \gets \text{sample from } p_\theta(\hat{x}_{t-1} | \hat{x}_t)$ \Comment{unconditional denoising}
    \State ${y}_{t-1} \gets \text{sample from}~\mathcal{N}({y}_{t-1}; \sqrt{\bar \alpha_{t-1}}y_{0},   (1 - \bar \alpha_{t-1})\boldsymbol{\mathrm{I}})$
    \State $\hat{x}_{t-1} \gets \hat{x}_{t-1} - \phi_N(\hat{x}_{t-1}) + \phi_N({y}_{t-1})$ \Comment{low frequency content constraint}
    \EndFor
\State $\hat{x}_{0} \gets ({x}_{\tau} - \sqrt{1-\bar{\alpha}_{\tau}}\boldsymbol{\epsilon}_{\theta}(x_{\tau}, \tau)) / \sqrt{\bar{\alpha}_{\tau}}$ \Comment{Estimate $x_0$ from $x_{\tau}$ directly}
\State $\hat{x}_0 \gets (\hat{x}_0)_{{\uparrow}_r}$
\State $x_0 \gets f(\hat{x}_0)$ \Comment{Run post-processing restoration model}
\State \Return $x_{0}$
\end{algorithmic}
\label{alg:sampling_dr2}
\end{algorithm}
\end{figure}


\clearpage

{\small
\bibliographystyle{ieee_fullname}
\bibliography{egbib}
}

\end{document}